\documentclass[10pt,onecolumn]{article}

\usepackage{graphicx}
\usepackage[colorlinks=true,  linkcolor=blue, citecolor=blue]{hyperref}
\usepackage{amsmath}
\usepackage{subcaption}
\usepackage{enumitem}
\usepackage[top=2.5cm,left=2cm,bottom=2.5cm,right=2cm]{geometry}

\usepackage{authblk}
\title{\bf Enhancing Evaluation Methods for Infrared Small-Target Detection in Real-world Scenarios}
\author[1]{Saed Moradi}
\author[1]{Alireza Memarmoghadam}
\author[1]{Payman Moallem\thanks{corresponding author: p\textunderscore moallem@eng.ui.ac.ir}}
\author[1]{Mohamad Farzan Sabahi}
\affil[1]{Department of Electrical Engineering, Faculty of Engineering, University of Isfahan, Isfahan, Iran}
\date{}                     
\begin{document}

\maketitle
\begin{abstract}
Infrared small target detection (IRSTD) poses a significant challenge in the field of computer vision. While substantial efforts have been made over the past two decades to improve the detection capabilities of IRSTD algorithms, there has been a lack of extensive investigation into the evaluation metrics used for assessing their performance. In this paper, we employ a systematic approach to address this issue by first evaluating the effectiveness of existing metrics and then proposing new metrics to overcome the limitations of conventional ones. To achieve this, we carefully analyze the necessary conditions for successful detection and identify the shortcomings of current evaluation metrics, including both pre-thresholding and post-thresholding metrics. We then introduce new metrics that are designed to align with the requirements of real-world systems. Furthermore, we utilize these newly proposed metrics to compare and evaluate the performance of five widely recognized small infrared target detection algorithms. The results demonstrate that the new metrics provide consistent and meaningful quantitative assessments, aligning with qualitative observations.

\textbf{Keywords:} Infrared small target detection; thresholding; pre-thresholding metrics; post-thresholding metrics
\end{abstract}
\section{Introduction}

Nowadays, infrared (IR) imaging has a wide range of application from medical \cite{chudecka2022use,moradi2022infrared}  and industrial diagnosis \cite{CHOUDHARY2021109196} to defense \cite{shahraki2021infrared} and remote sensing \cite{raza2021ir}. Generally, processing IR images is a challenging task \cite{zhang2022isnet} due to the specifications of IR imaging. Among all the aforementioned applications,  IR small target detection (IRSTD) is a highly challenging research field  because:
\begin{itemize}
\item Since the IR small targets are far from the imaging device, the target has low local contrast and appears as a dim spot in the image plane \cite{shahraki2022infrared}.
\item The small target in IR images typically occupies handful of  pixels \cite{moradi2020fast}. Thus, the region of interest (ROI) does not represent distinguished features.
\item The edges of the small target are blurred due to atmospheric thermal fields \cite{khaledian2014new}. Therefore, there are not a clear boundary between background area and target pixels. 
\end{itemize}

\begin{figure}[h]
\centering
\includegraphics[width=0.9\textwidth]{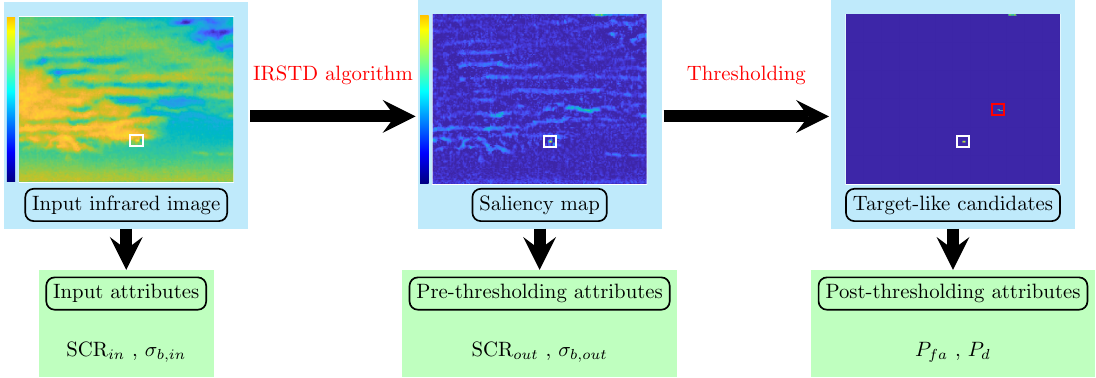}
\caption{The block diagram of a typical IRSTD pipline}
\label{fig:IRST_block_diagram}
\end{figure}

The block diagram of a typical IRSTD pipeline is illustrated in \autoref{fig:IRST_block_diagram}. As shown in the figure, the input IR image is first process by the IRSTD algorithm to create a saliency map. Note that, while the IRSTD algorithm may refer to the end-to-end IR image processing pipeline, here, the process of construction of saliency map from input IR image is called IRSTD algorithm. The goal is to suppress the background area and enhance target pixels. An ideal saliency map should eliminate the background intensities and only preserve the target area. After saliency map reconstruction, a thresholding strategy is chosen to be applied on the saliency map. Then, true (logical one) pixels in the resulting binary image is considered as target-like objects.   

Considering the pipeline in the  \autoref{fig:IRST_block_diagram}, specific attributes are defined for images in pipeline. 
	In the IRSTD terminology, the input image can be described using two attributes:
	\begin{itemize}
	
	\item $\sigma_{b,in}$: This represents the standard deviation of background pixels in the input image, which is directly related to the complexity of the background. A smaller value of $\sigma_{b,in}$ indicates a smoother background, while a larger value indicates a more complex background.	
	\item SCR$_{in}$: This acronym stands for the signal-to-clutter ratio in the input image. SCR is defined as $\frac{\mu_t-\mu_b}{\sigma_b}$, where $\mu_t$, $\mu_b$, and $\sigma_b$ denote the mean value of target pixels, mean value of local background pixels, and standard deviation of the local background, respectively.
		\end{itemize} 

Same argument is valid for saliency map (The processed image by IRSTD algorithm). Thus, just like the input attributes, $\sigma_{b,out}$ and SCR$_{out}$ represents the background complexity and the signal to clutter ration in the saliency map. It is clear that for a typical IRSTD pipeline:
\begin{equation}
\sigma_{b,out} < \sigma_{b,in} \quad \text{and} \quad \text{SCR}_{out} > \text{SCR}_{in}
\label{eq:enhancementparam}
\end{equation}

According to \autoref{eq:enhancementparam}, two performance metrics are defined for evaluation of IRSTD algorithms: Background suppression factor (BSF) and signal to clutter ration gain (SCRG) which are defined as follows \cite{bai2022floating}:
\begin{equation}
\text{BSF}=\frac{\sigma_{b,in}}{\sigma_{b,out}} \quad , \quad \text{SCRG}=\frac{\text{SCR}_{out}}{\text{SCR}_{in}}
\label{eq:bsfscrg}
\end{equation}

Based on \autoref{eq:enhancementparam} and  \autoref{eq:bsfscrg}, larger values for both SCRG and BSF are desired. Note that, in case of evaluation of different IRSTD algorithms, since the input images are the same for all baseline algorithms, $\text{SCR}_{out}$ and $\frac{1}{\sigma_{b,out}}$ can be used as performance metrics, as well.

The IRSTD algorithms are well-studied in the literature. Mainly, these algorithms can be categorized based on filtering method,  contrast measure calculation, and data structure decomposition \cite{zhao2022single}. 

The filtering based methods are divided into two sub-categories. The first one is the spatial domain filtering, in which, the input infrared image is processed using local kernels to enhance the target area. Max-mean \cite{deshpande1999max}, max-median \cite{deshpande1999max}, bilateral filtering \cite{bae2014spatial}, morphological operators \cite{zhu2020balanced}, two dimensional least mean square \cite{han2019local} are some instances of this sub-category. The second one refers to processing in the transformation domain. In these techniques, the input image is transformed to a desired transformation space like as frequency \cite{wang2017infrared} and  wavelet \cite{wang2012detecting} domains. Then, after processing the transformed information, the inverse transform is applied to recover true targets.

Methods based on human visual systems (HVS) which lead to local contrast-based mechanism   has received researchers' attention during last few years. These methods outperform filter-based methods in terms of SCRG and BSF. However, they usually have higher computational complexity compared to filter-based ones. Generally, local contrast can be constructed in either difference or ratio forms. Difference local contrast like as Laplacian of Gaussian (LoG) \cite{kim2012scale}, difference of Gaussian (DoG) \cite{wang2012infrared}, improved difference of Gabor \cite{han2016infrared}, center-surround difference measure \cite{xie2014small}, and local difference adaptive measure \cite{li2019small}. Unlike the difference form local measures, ratio-form local measures utilize enhancement factor which is the ration between the center cell and surrounding ones. Local contrast measure (LCM) \cite{chen2013local}, improved local contrast measure (ILCM) \cite{han2014robust}, relative local contrast measure (RLCM) \cite{han2018infrared}, Tri-Layer local constrast method (TLLCM) \cite{han2019localb}, novel local contrast descriptor (NLCD) \cite{qin2019infrared},  absolute directional mean difference (ADMD) \cite{moradi2020fast}, and  weighted strengthened local contrast measure (WSLCM) \cite{han2020infrared} are the most effective IRSTDs in the literature. There  is also a combined local measure which benefits from both difference and ratio    from of local contrast measure \cite{han2022ratio}. 

Data structure decomposition-based methods are also a newly introduced class of IRSTDs.  Sparse and low-rank matrices decomposition is the principal of these class of IRSTDs. Infrared patch image (IPI) model \cite{gao2013infrared}, weighted infrared patch image (WIPI) model \cite{dai2016infrared}, non-negative infrared patch image model based on partial sum minimization of singular values (NIPPS) \cite{dai2017non}, nonconvex rank approximation minimization (NRAM) \cite{zhang2018infrared}, and nonconvex optimization with an L$_p$ norm constraint (NOLC) \cite{zhang2019infrared} are the recent efforts of IR image decomposition-based approach.

	Deep learning-based methods, unlike manually designed operators, have the advantage of leveraging the abundance of data available to effectively learn the target model. The work in \cite{wang2019miss} involved training generators using a generative adversarial network to address miss detection and false alarm aspects. ACM \cite{dai2021asymmetric} devised feature fusion modules within the encoder-decoder framework to enhance the representation of low and deep semantic features, resulting in a more effective feature representation. ALCNet \cite{dai2021attentional} employs shift operations on semantic tensors to simulate local contrast, aiming to extract specific local information about the target. 
	In \cite{li2022dense}, the authors introduce a dense nested interactive module (DNIM) to maintain the information of small targets in deep layers through progressive interaction. Additionally, a cascaded channel and spatial attention module (CSAM) is proposed to adaptively enhance multi-level features, enabling effective incorporation and exploitation of contextual information.
		In \cite{ying2023mapping}, the authors proposed a label evolution framework called Label Evolution with Single Point Supervision (LESPS) to address the labor-intensive nature of fully supervised training for infrared small target detection in CNNs. By leveraging the phenomenon of "mapping degeneration," where CNNs initially learn to segment pixels near the targets before converging to predict ground truth point labels, LESP enables progressive expansion of the point labels using intermediate predictions.

When a saliency map is generated prior to the detection task, the  dream is to obtain larger BSF and SCRG values . However, having larger BSF and SCRG does not guarantee a successful detection. A high performance IRSTD algorithm should be followed by a proper thresholding strategy to detect real targets and eliminate false responses. This is why there are two more performance metrics after applying the thresholding operation to the saliency map. These two metrics which demonstrate the ability of detection true targets and eliminating false responses are called probability of detection $P_d$ and probability of false alarms $P_{fa}$, respectively.  In contrast to BSF and SCRG which are measurable before applying the threshold (This is why we call them pre-thresholding attributes), these two metrics are measured on binary images and therefore we call them post-thresholding attributes (\autoref{fig:IRST_block_diagram}).   

As mentioned in the previous paragraph, for a successful detection, both high performance IRSTD algorithm as well as the proper thresholding strategy are required. Regardless of effectiveness of the IRSTD algorithm, improper thresholding will leads to missing true targets and having false responses which could be disaster for a practical system. Hence, in this paper, after investigating various thresholding strategies, the best methods for applying threshold to the saliency map is presented. Then, current pre-thresholding as well as the post-thresholding metrics are investigated, and some new metrics which are aligned with practical considerations are proposed. The rest of this paper is organized as follows: in the next section, the role of thresholding in practical systems is deeply investigated. Then, in section 3, current pre-thresholding metrics are reviewed. After demonstrating their shortages, modified metrics are proposed for IRSTD performance  evaluation. In section 4, same process is performed for post-thresholding metrics. In section 5, The newly proposed metrics are used for performance comparison of common IRSTD algorithms. Finally, the paper is concluded in section 6.     

\section{The onus of thresholding on the overall performance }
\begin{figure}[t!]
   	\centering
   	\subfloat[]{\includegraphics[height=1.5in]{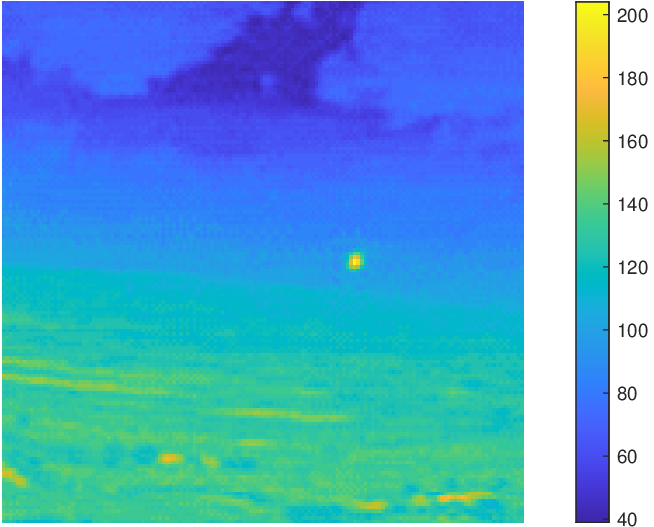}%
   		\label{fig:indr}}
   		   	~~
   	\subfloat[]{\includegraphics[height=1.5in]{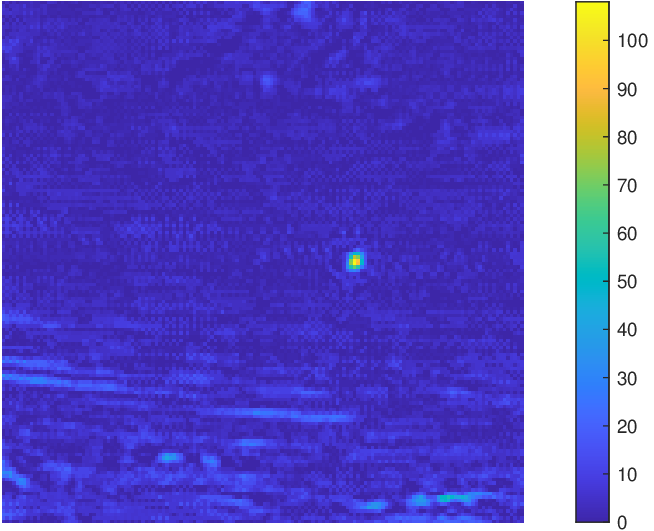}%
   		\label{fig:tophatdr}}
   	~~
   	\subfloat[]{\includegraphics[height=1.5in]{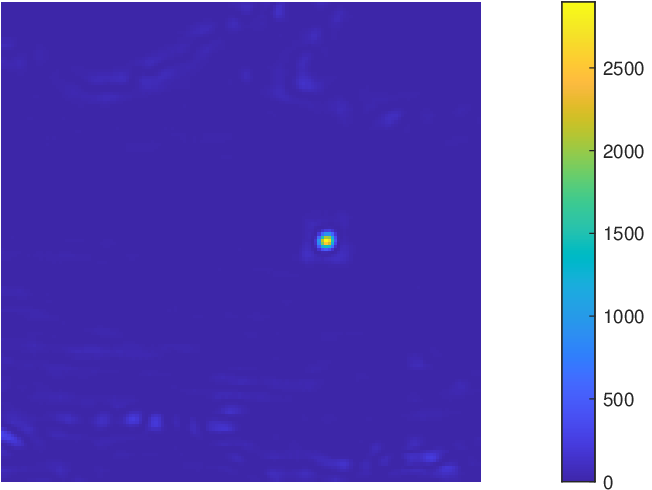}%
   		\label{fig:aagddr}}
   	
   	\caption{Variable dynamic range in saliency map. a) Original infrared image. b) filtering result using TopHat algorithm \cite{zhu2020balanced}, c) filtering result using AAGD \cite{moradi2018false} algorithm. The dynamic range of the saliency map might be different than input infrared image depending on the applied IRSTD.}
   	\label{fig:inoutDR}
   \end{figure}

\begin{figure}[t!]
   	\centering
   	\subfloat[]{\includegraphics[width=1.4in]{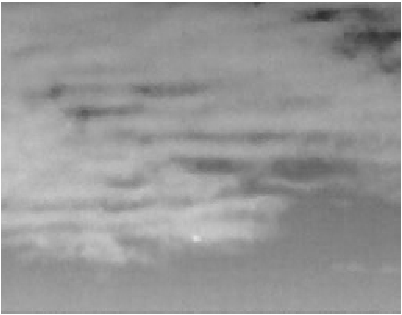}%
   		\label{fig:originaltt1}}
   	~~
   	\subfloat[]{\includegraphics[width=1.4in]{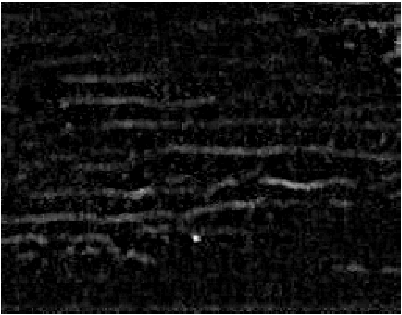}%
   		\label{fig:tophattt1}}
   	~~
   	\subfloat[]{\includegraphics[width=1.4in]{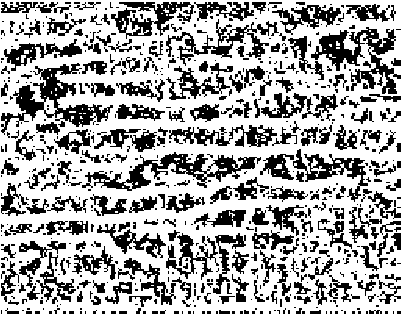}%
   		\label{fig:otsutt1}}
   	\\
   	\subfloat[]{\includegraphics[width=1.4in]{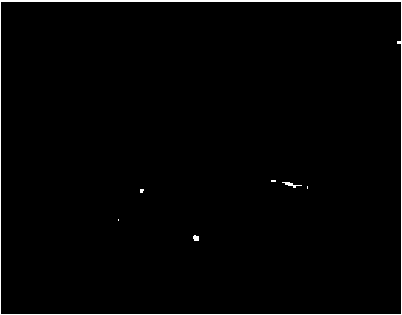}%
   		\label{fig:autott1}}
   	~~
   	\subfloat[]{\includegraphics[width=1.4in]{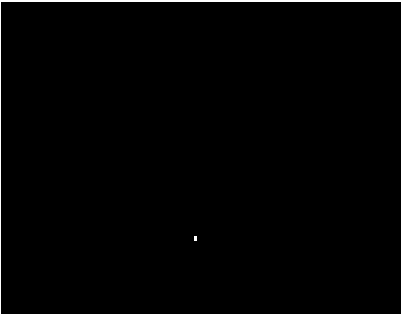}%
   		\label{fig:manualtt1}}
   	
   	\caption{The automatic thresholding results. a) Original infrared image. b) Top-hat filtering result \cite{zhu2020balanced}. c) Otsu's thresholding result $(T=0.48)$. d) automatic thresholding using average values of background and target classes $(T=19)$. e) Manual thresholding $(T=29)$.}
   	\label{fig:thresh_test}
   \end{figure}

After performing target enhancement and clutter suppression procedure (saliency map construction), the filtered IR image should be converted to binary one using  thresholding operation that can be applied in different forms (i.e manual, automatic, local, global). Since the target detection problem only consists of two different classes namely as target and background clutter,  single-level thresholding is a satisfactory option for this purpose. The simplest method to achieve the classification goal, is to apply a global threshold  $T$:

   \begin{equation}
   g(x,y) = 
   \begin{cases} 
   1 & \quad \quad \quad  f(x,y) > T \\
   0 & \quad \quad \quad  f(x,y) \leq T
   \end{cases}
   \end{equation}
where,  $f(x,y)$ and $g(x,y)$  stand for the saliency map and binary image,respectively. The most challenging part of global thresholding operation is how to set an effective threshold value $T$. Since target detection systems continuously scan the  environment, human operator cannot be helpful to choose the optimum threshold value. The most simplest way to do this is to choose a unique threshold value based on experiments for all incoming image frames. However, when the dynamic range of the filtered image is not equal to the dynamic range of the input images, the false-alarm rate or the miss-rate will increase drastically. \autoref{fig:inoutDR} shows the change in the dynamic range of filtered images (saliency maps) using Tophat and AAGD IRSTDs. As shown in the figure; the output dynamic range directly depends on the applied IRSTD.  Therefore, the thresholding procedure should be performed in an automated manner.   There are various  automatic image thresholding algorithms in the literature. The Otsu's method is one of the widely used one \cite{otsu1979threshold}. In this method the global threshold value is chosen in a way, to maximize inter-class variance. When both foreground (Target) and background classes include considerable number of pixels, and the image histogram is bimodal (i.e. there is a deep valley between two peaks in the image histogram), the Otsu's method works very well in object segmentation problems. However, when the target area is too small compared to the background area,  which is always occurred in incoming infrared target detection problems, the segmentation result of Otsu's method is inaccurate \autoref{fig:thresh_test}. Another widely used automatic thresholding is presented in \cite{gonzalez2011digital}, where the followings  are performed to obtain the desired threshold value:
\begin{enumerate}[label=\roman*)]
\item The gray image is segmented into two classes using threshold value equal to global mean of the image ($T=\mu_G$).
\item The average values of the background and target  are calculated $(\mu_B,~ \mu_T)$.
\item  The new threshold level is calculated $\left(T_{new}=\frac{1}{2}\left( \mu_T + \mu_B \right)\right) $.
\item While $(T_{new}-T_{old}>\epsilon)$, steps (ii) and  (iii) are recursively repeated.
\end{enumerate}  
   
\begin{figure}[t!]
   	\centering
   	\subfloat[]{\includegraphics[width=1.4in]{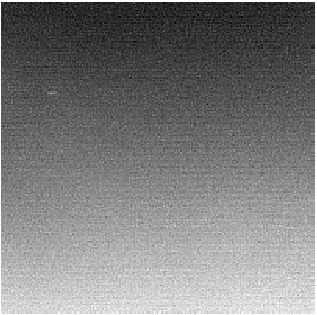}%
   		\label{fig:originaltt2}}
   	~~
   	\subfloat[]{\includegraphics[width=1.4in]{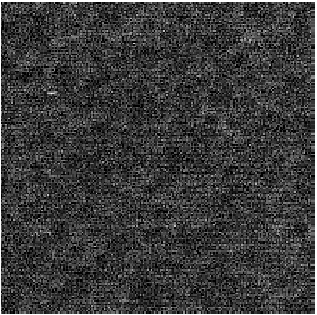}%
   		\label{fig:tophattt2}}
   	~~
   	\subfloat[]{\includegraphics[width=1.4in]{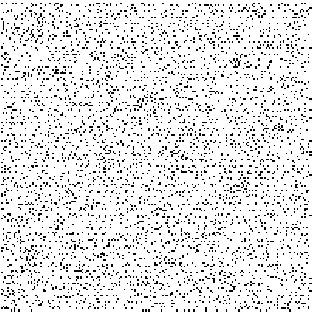}%
   		\label{fig:otsutt2}}
   	\\
   	\subfloat[]{\includegraphics[width=1.4in]{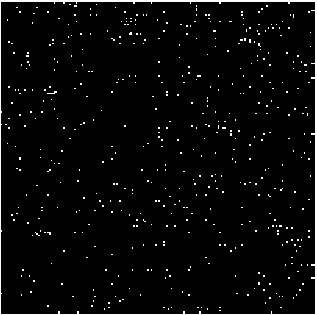}%
   		\label{fig:autott2}}
   	~~
   	\subfloat[]{\includegraphics[width=1.4in]{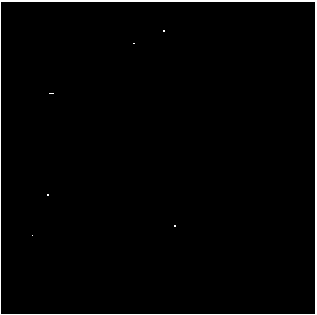}%
   		\label{fig:manualtt2}}
   	
   	\caption{The automatic thresholding results. a) Original noisy infrared image. b) Top-hat filtering result. c) Otsu's thresholding result $(T=0.5)$. d) automatic thresholding using average values of background and target classes $(T=7)$. e) Manual thresholding $(T=10)$.}
   	\label{fig:thresh_test2}
   \end{figure}

\begin{figure}[t!]
   	\centering
   	\subfloat[]{\includegraphics[width=1.4in]{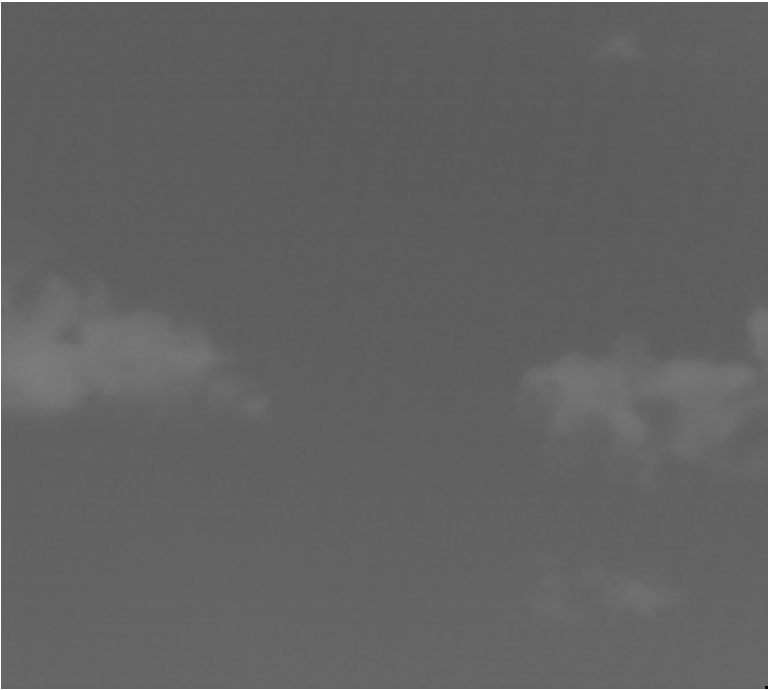}%
   		\label{fig:original_without_target}}
   	~~
   	\subfloat[]{\includegraphics[width=1.4in]{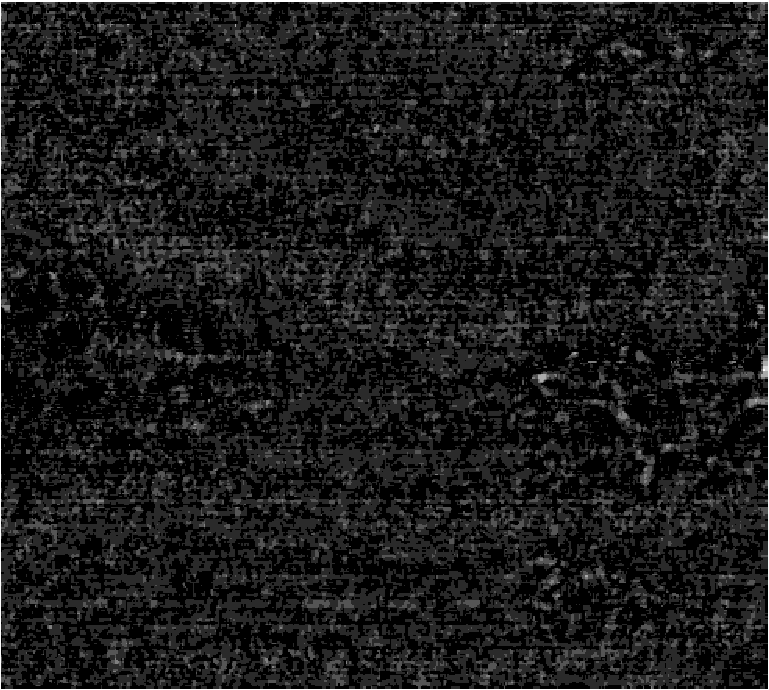}%
   		\label{fig:tophat_without_target}}
   	~~
   	\subfloat[]{\includegraphics[width=1.4in]{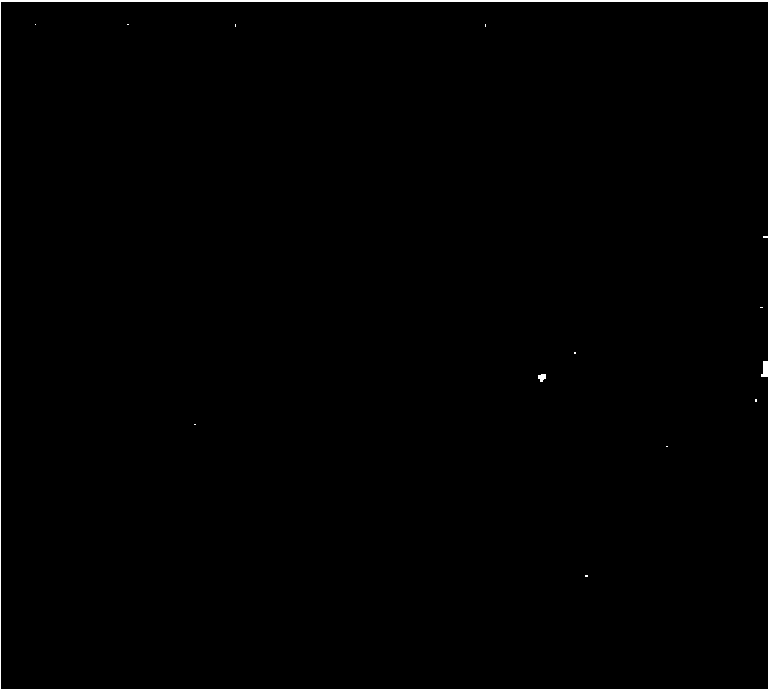}%
   		\label{fig:thresh_without_tar}}
   	\caption{Drawback of  automatic thresholding  in scenarios with no targets. a) original infrared image which does not contain small target. b) the result of Top-Hat filtering. c) automatic thresholding results. }
   	\label{fig:thresh_without_target}
   \end{figure}

   When the background noise is not strong,   this automatic thresholding operation shows good performance for final target detection (\autoref{fig:autott1}). However, in strong noisy scenarios, the performance of this algorithm is degraded significantly (\autoref{fig:autott2}), which in turn, increases the false responses.  Moreover, when infrared scenario does not contain any small target, these histogram-based automatic thresholding methods always return incorrect responses in non-target areas (\autoref{fig:thresh_without_target}). 
   
  Statistics-based image thresholding is the most effective thresholding strategy for small target detection which can be applied in both local and global manners \cite{gonzalez2011digital,qin2016effective}. Statistics-based global and local thresholding are expressed in \autoref{eq:adaptive_global_thresh} and \autoref{eq:adaptive_local_thresh}, respectively.
    \begin{equation}
   T=\mu_G+k_G\times \sigma_G
   \label{eq:adaptive_global_thresh}
   \end{equation}  
        \begin{equation}
   T(x,y)=\mu(x,y)+k_L\times \sigma(x,y)
      \label{eq:adaptive_local_thresh}
   \end{equation}
   where, $\mu_G$, $\sigma_G$, $\mu(x,y)$, $\sigma(x,y)$,  $k_G$, and $k_L$ indicate global mean of the image, global standard deviation of the image, local mean around $(x,y)$ position, local standard deviation around $(x,y)$ position, control parameter of global thresholding and  local thresholding, respectively.
   
Global thresholding is a simple operation with low computational complexity. However, in multi-target scenarios, some  targets  may be missed. Local thresholding can  detect all targets. Since local mean and standard deviation should be calculated for each pixel in the gray image, the local statistics-based thresholding has higher computational complexity compared to the global one.   Generally speaking, using statistics-based thresholding  has the following advantages:
  
\begin{itemize}
\item It can work with any gray-level dynamic ranges.
\item The control parameter $(k)$ can be determined by experiments to achieve reasonable false-alarm rate.
\item  The last but not the least, it is very effective for scenarios with no targets. 
\end{itemize}

 \section{Pre-thresholding evaluation}  
	Typically, infrared small target detectors (IRSTDs) incorporate a pre-processing stage that includes a module for removing single-pixel high-brightness noise (SPHBN). In this case, as the SPHBN has already been eliminated, the success of the detection process relies on accurately recognizing even a single pixel within the target area and the exact boundary extraction of the target area is not important at all. Therefore, a proper evaluation metric should support this argument.
 
\begin{figure}[t!]
   	\centering
   	\subfloat[]{\includegraphics[width=1.4in]{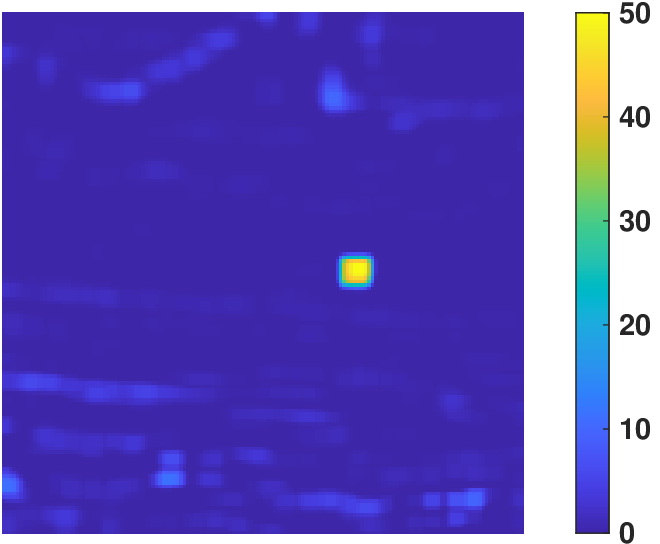}%
   		\label{fig:scrDout1}}
   	~~
   	\subfloat[]{\includegraphics[width=1.4in]{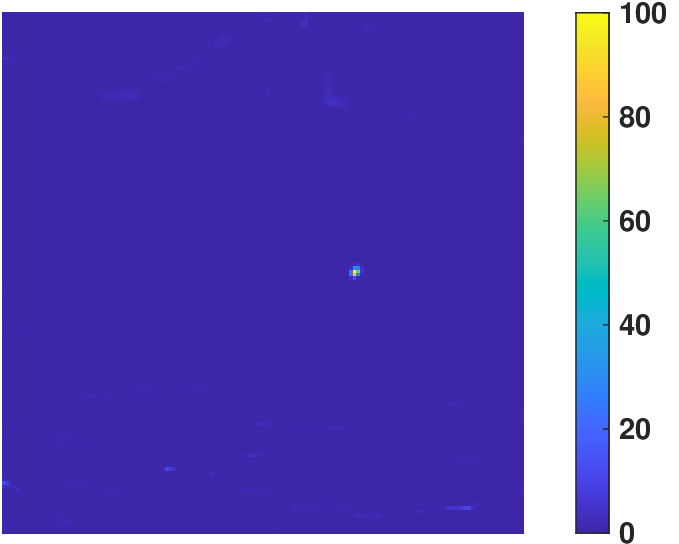}%
   		\label{fig:scrDout2}}
   ~~
   	\subfloat[]{\includegraphics[width=1.4in]{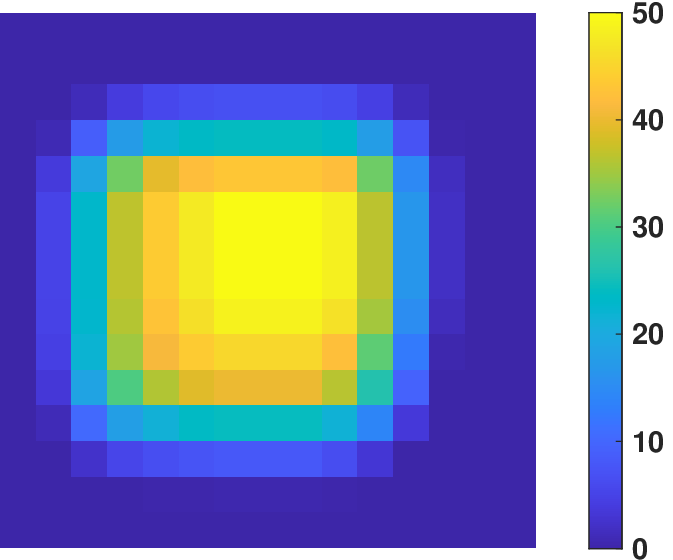}%
   		\label{fig:scrDout1_zoomed}}
   		   	\\
   	\subfloat[]{\includegraphics[width=1.4in]{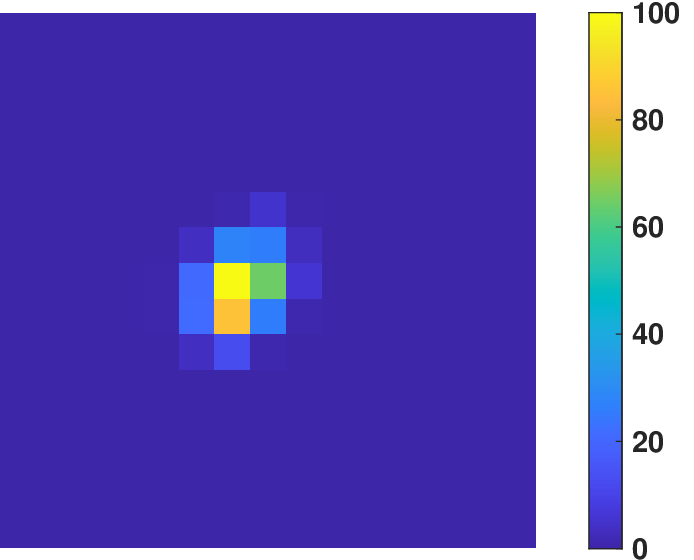}%
   		\label{fig:scrDout2_zoomed}}
   		   ~~
   	\subfloat[]{\includegraphics[width=1.4in]{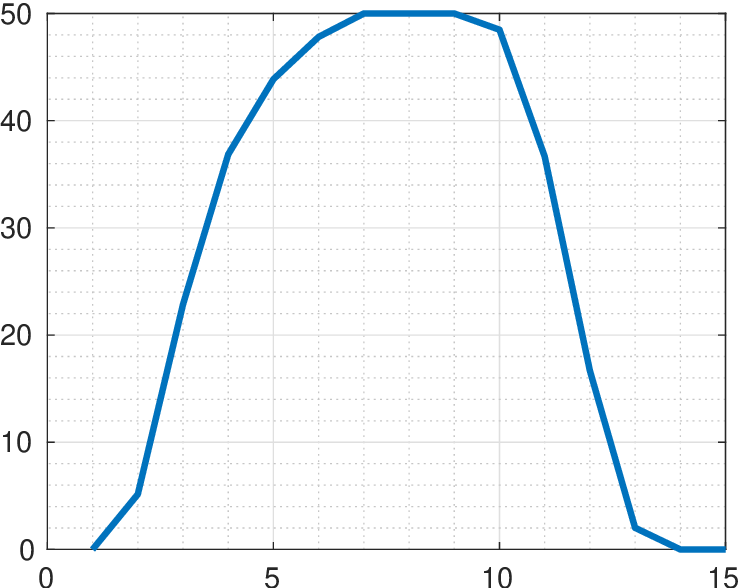}%
   		\label{fig:oneDcrossout1}}
   		   	~~
   	\subfloat[]{\includegraphics[width=1.4in]{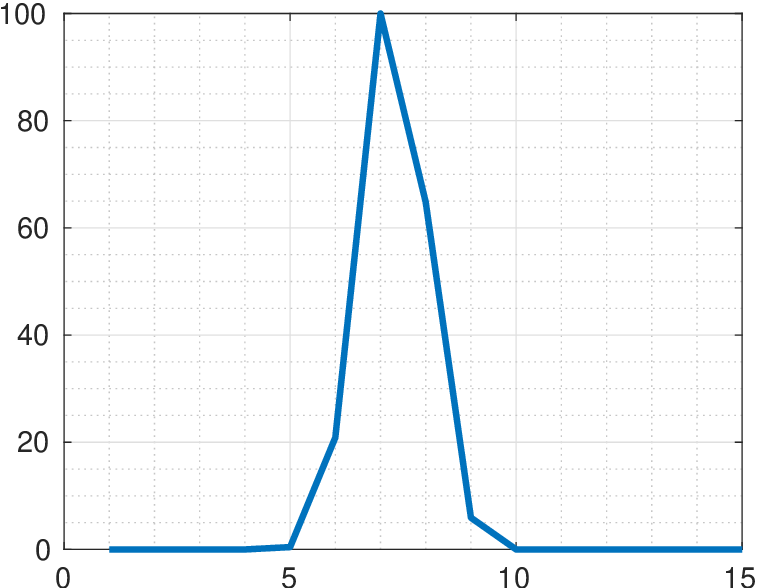}%
   		\label{fig:oneDcrossout2}}
   		\\
   		   	\subfloat[]{\includegraphics[width=0.85\textwidth]{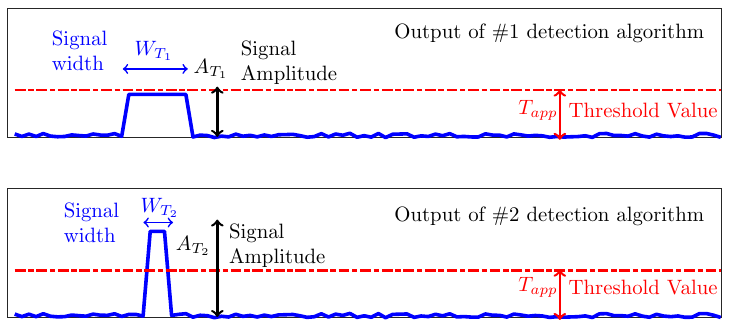}%
   		\label{fig:SCR_disadvOned}}
   	\caption{A simple scenario to demonstrate the drawback of  common SCR metric.   a) target area in the saliency map of the IRSTD $\# 1$, b) target area in the saliency map of the IRSTD $\# 2$, c) $1$D plot of target cross-section in \subref{fig:scrDout1_zoomed}, d)  c) $1$D plot of target cross-section in \subref{fig:scrDout2_zoomed}, e) simplified $1$D representation of target area in both IRSTD $\# 1$ and $\# 2$.}
   	\label{fig:SCR_disadv}
   \end{figure}

 Signal to clutter ratio (SCR) is one of  pre-thresholding metrics which shows the target enhancement ability of an IRSTD, which is defined as:
       \begin{equation}
   SCR=\frac{\mu_T-\mu_b}{\sigma_b},
   \label{eq:rec_scr}
   \end{equation}
where, $\mu_T$, $\mu_b$, and $\sigma_b$ denote average intensity of the target area, average intensity and standard deviation of its local surrounding background, respectively.  While this evaluation measure is generally accepted in the literature, it can not correctly reflect the target enhancement capability of an IRSTD. To better understanding, a simple scenario is provided here (\autoref{fig:SCR_disadv}).

Two different saliency maps are demonstrated in \autoref{fig:scrDout1} and \autoref{fig:scrDout2}. \autoref{fig:scrDout1} shows the result of applying AAGD algorithm \cite{moradi2018false}  with $9\times 9$ internal window followed by a morphological dilation with a $3\times 3$ square-shape structural element. As depicted in the figure; the target area is relatively enhanced while there are some remaining background clutter. Compared to the \autoref{fig:scrDout1}, the second IRSTD which  again is an AAGD \autoref{fig:scrDout2} algorithm with $3\times 3$ internal window followed by a morphological erosion with a $3\times 3$ square-shape structural element, shows better target enhancement and background suppression. As shown in \autoref{fig:scrDout1_zoomed} and \autoref{fig:scrDout2_zoomed}, the signal amplitude for the IRSTD $\# 2$ is almost twice as the one in the IRSTD $\# 1$, which means in higher threshold values the target will be detected corectly in the second one, while in the first one the target will be missed. One dimensional ($1$D) cross-section of target area in both saliency maps are shown in \autoref{fig:oneDcrossout1} and \autoref{fig:oneDcrossout2}, respectively. To simplify the scenario, let's approximate $1$D cross-section of the target area with closest square signal. The result is simplified and shown in \autoref{fig:SCR_disadvOned}. As shown in the figure:
\begin{equation}
A_{T_1}=\frac{A_{T_2}}{2} \quad , \quad W_{T_1}=3\times W_{T_2}
\end{equation}
 where, $A_{T_1}$, $A_{T_2}$ denote the target amplitude in the output of IRSTD $\#1$ and $\#2$. Also, $ W_{T_1}$, $ W_{T_2}$ show  the target width (extension) in the output of IRSTD $\#1$ and $\#2$, respectively.

Based on SCR formulation (\autoref{eq:rec_scr}), and  simply considering zero-mean background signal, the following relationship can be easily derived:
 \begin{equation}
SCR_1=\frac{3}{2}\times SCR_2
\label{eq:scr1scr2}
\end{equation}
which implies  that the target detection ability of the IRSTD $\#1$  is $50\%$ more than that of IRSTD $\#1$. However, by applying a global  threshold level at $T_{app}$, the $\#1$ algorithms does not detect the true target (\autoref{fig:SCR_disadv}). It can be  clearly seen that the $\#2$ algorithm  can detect the true target at the same threshold level. In order to address this issue when global thresholding is final choice in practical system,  the SCR formulation should be modified as:
       \begin{equation}
 \boxed{  SCR_{global}=\frac{\max_T-\mu_G}{\sigma_G}},
   \label{eq:rec_scr_new}
   \end{equation}
where, $\max_T$ denotes the maximum gray value of the target area. According to \autoref{eq:adaptive_global_thresh},  the maximum acceptable control parameter is equal to newly defined SCR metric:
        \begin{equation}
   k_{G_{\max}}=SCR_{global}.
   \label{eq:rec_scr_new_kup}
   \end{equation}
There are two important points regarding the \autoref{eq:rec_scr_new_kup}:
\begin{enumerate}
\item Common SCR metric is not able to correctly reflect the target detection ability. The pre-thresholding evaluation should be performed in a global manner on the saliency maps.
\item The thresholding operation should be consistent with pre-thresholding evaluation metrics. For instance, in our case, the global statistics-based thresholding is the right choice.  
\end{enumerate}

So far, it is demonstrated that the global thresholding is the right one to be applied on the saliency map. In the next subsection, we demonstrate the drawback of the local statistics-based thresholding.
\subsection{Drawback of common local thresholding}
Now, let consider the case that local thresholding is supposed to be applied on the saliency map. According to \autoref{fig:SCR_disadv}, the local mean around target region can be calculated as follows:
   \begin{equation}
   \mu(x)=\frac{AW}{n}
   \label{eq:mu_pulse}
   \end{equation}
  where, $A$, $W$, $x$, and  $n$ stand for the target amplitude,  width (spatial extension), the current index and number of samples in local neighborhood ($n>W$).

The local standard deviation can be calculated as:
   \begin{equation}
   \sigma(x)=\sqrt{\frac{1}{n}\sum_{i=1}^{n}\left( y(i)-\mu(x) \right)^2}
   \end{equation}
Where $y(i)$ and $\mu(x)$ denote the saliency map samples and local mean, respectively. Since the detection algorithm is supposed to suppress background clutter, for the sake of simplicity, we can assume that the saliency map samples out of the target region are equal to zero. Then:
%
%
%
       \begin{equation}
   \sigma(x)=\frac{A}{n}\left(\sqrt{ nW-W^2 }\right)
   \end{equation}  
   
   The local thresholding (\autoref{eq:adaptive_local_thresh}), can be rewritten as:
      \begin{equation}
   T(x)= \mu(x)+k_L\times \sigma(x)=\frac{AW}{n}+ \frac{k_LA}{n}\left(\sqrt{ nW-W^2 }\right)
   \end{equation} 

The detection process is established correctly for the threshold values  lower than target amplitude ($T(x)<A$). Therefore, for a successful target detection the following condition should be met:
   \begin{equation}
   \begin{split}
   & \mu(x)+k_L\times \sigma(x) <A \\
   &\Rightarrow\frac{AW}{n}+ \frac{k_LA}{n}\left(\sqrt{ nW-W^2 }\right)<A
   \end{split} 
   \end{equation}
   
\begin{figure}[t!]
   	\centering
   	\includegraphics[width=2.8in]{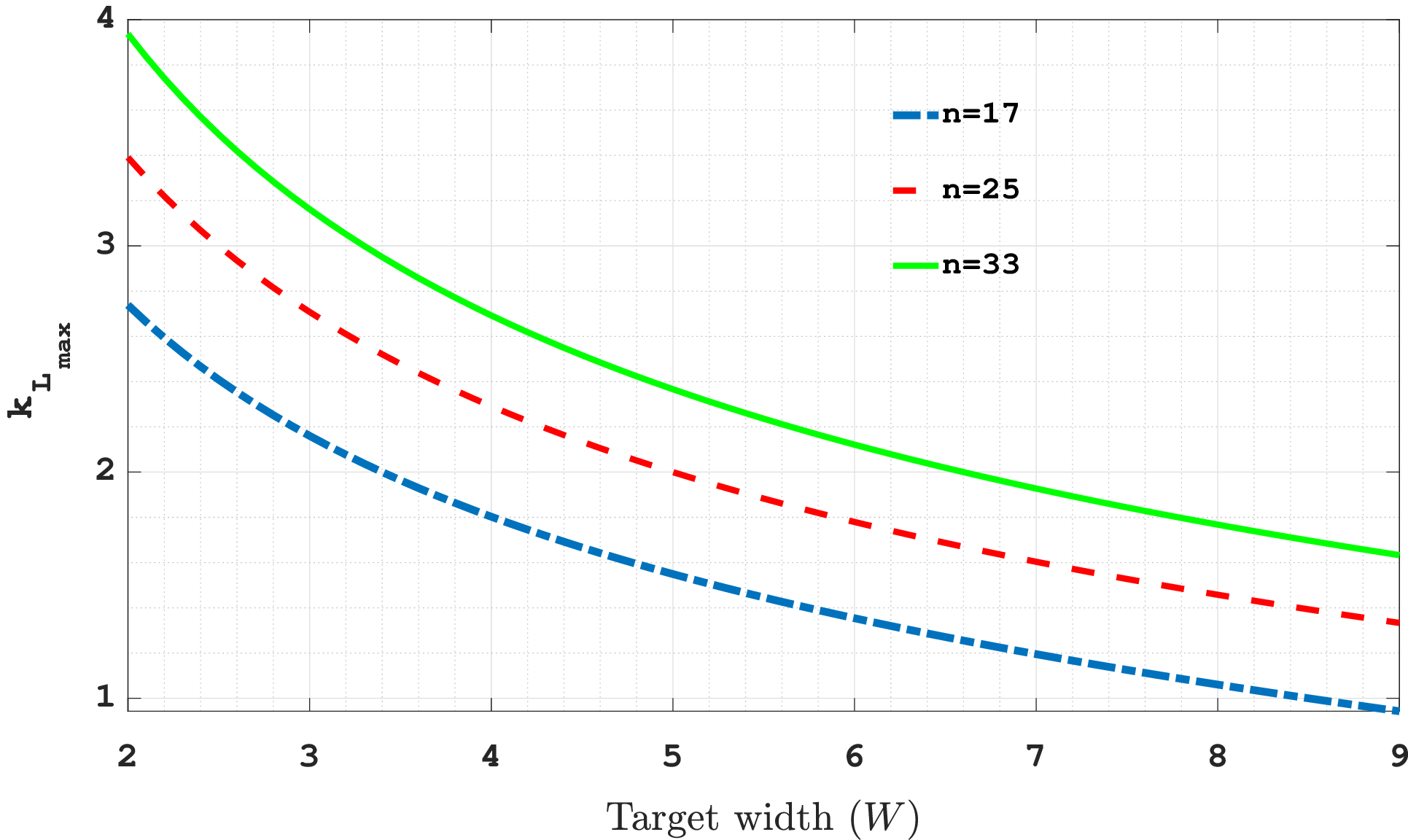}%
   	\caption{$k_{L_{\max}}$ versus target spatial extension $W$ (target width in $1$D case) }
   	\label{fig:kup_Vs_w}
   \end{figure}

   The upper bound for control parameter ($k_{L_{\max}}$) to detect the target accurately can be find as follows:
       \begin{equation}
   \begin{split}
   &\frac{AW}{n}+ \frac{k_{L_{\max}}A}{n}\left(\sqrt{ nW-W^2 }\right)=A \\
   &\Rightarrow k_{L_{\max}}=\sqrt{\frac{n-W}{W}}
   \end{split}
   \label{eq:kup_local}
   \end{equation}  
   \autoref{fig:kup_Vs_w} shows the upper bound of control parameter versus target width. As shown in the figure, the maximum control parameter to detect target correctly using local thresholding decreases as the target width increases. $k_{L_{\max}}$ takes its maximum value when the target width is equal to one pixel ($k_{L_{\max}}=\sqrt{n-1}$ for $W=1$). This result is quite consistent with the fact that the most effective target detection algorithm should suppress all background region and only returns a single pixel (Target centroid). 
    
         \begin{figure}[t!]
   	\centering
   	\subfloat[]{\includegraphics[width=2.8in]{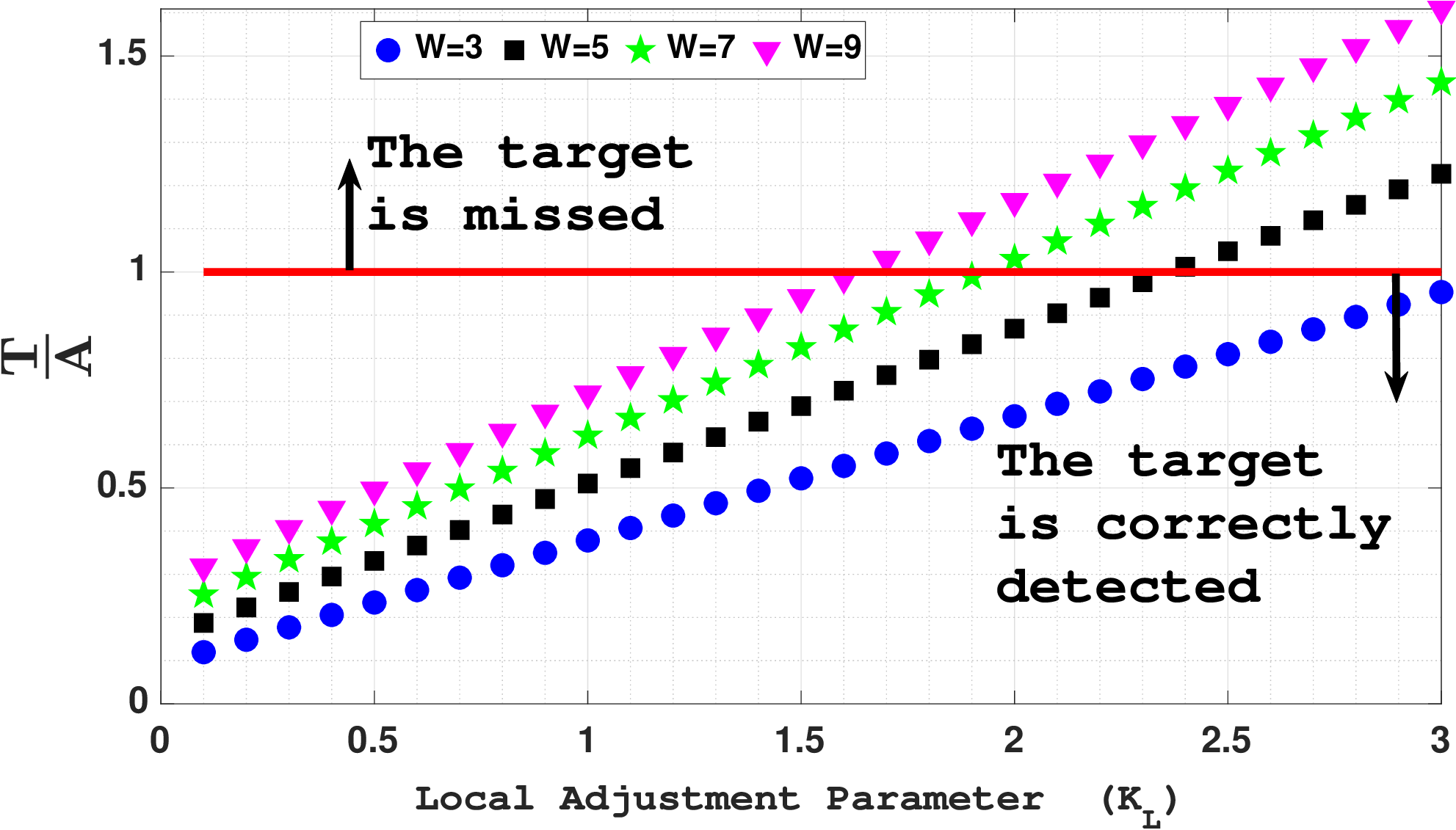}%
   		\label{fig:thresh_Vs_w33}}
   	~
   	\subfloat[]{\includegraphics[width=2.7in]{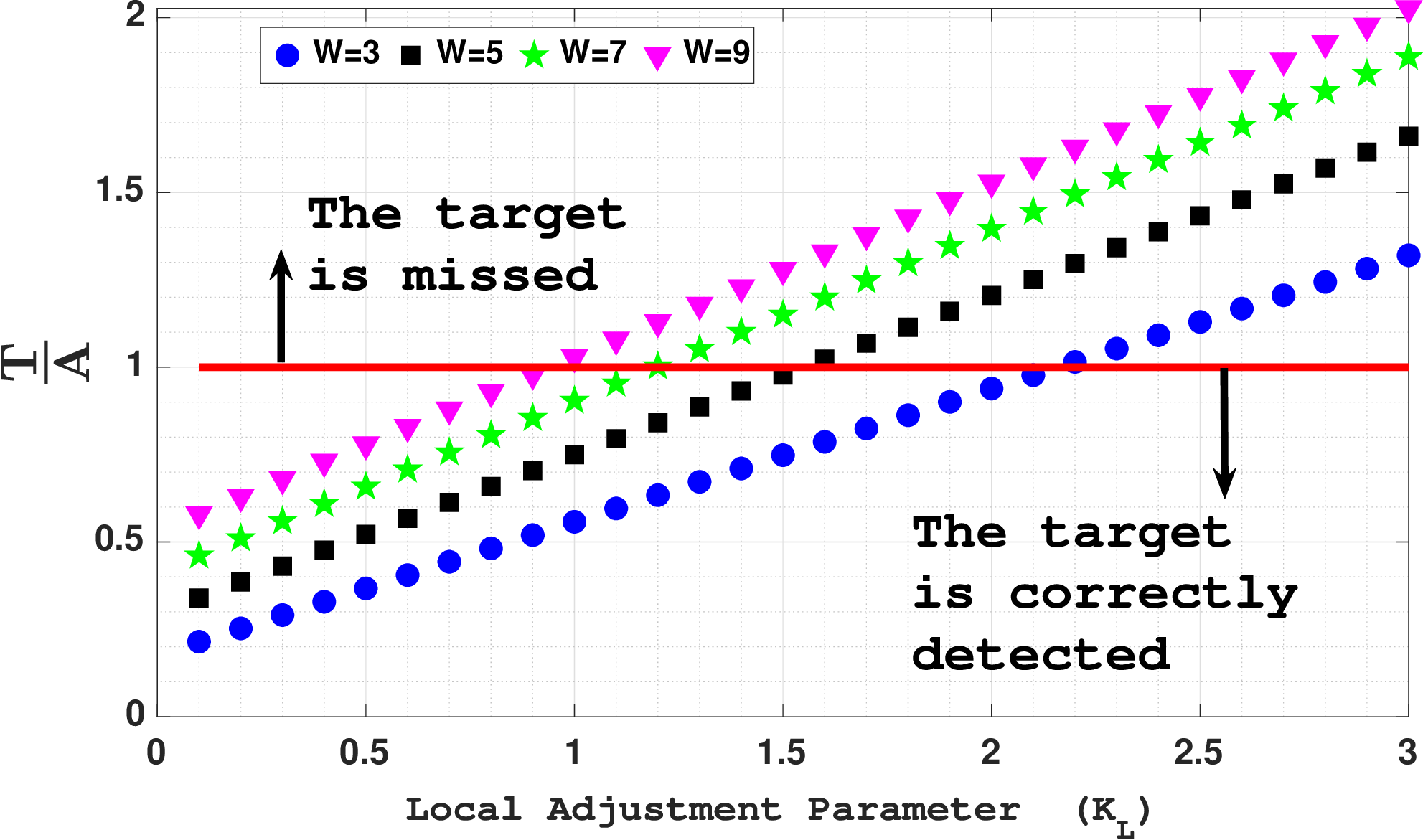}%
   		\label{fig:thresh_Vs_w17}}

   	\caption{Local threshold  value normalized to target amplitude ($\frac{T}{A}$) versus different control parameter. a) $n=33$, b) $n=17$. }
   	\label{fig:thresh_Vs_w}
   \end{figure}

\begin{figure}[t!]
   	\centering
   	\subfloat[]{\includegraphics[width=1.8in]{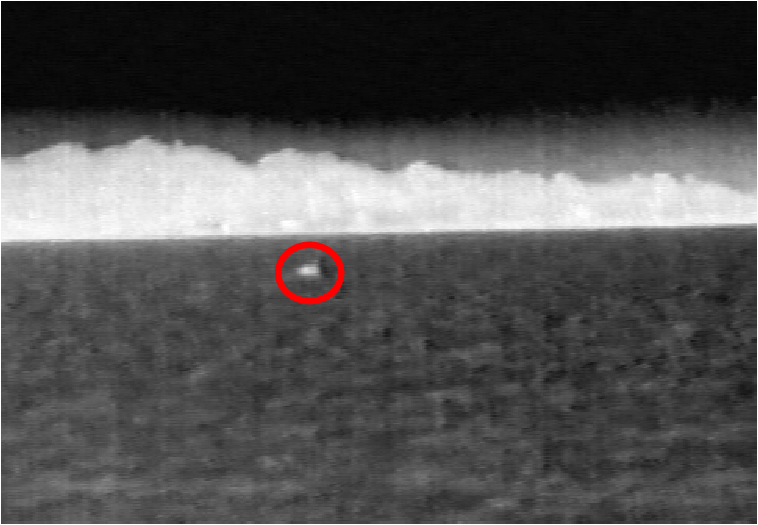}%
   		\label{fig:ldisdv_orig}}
   	~
   	\subfloat[]{\includegraphics[width=1.8in]{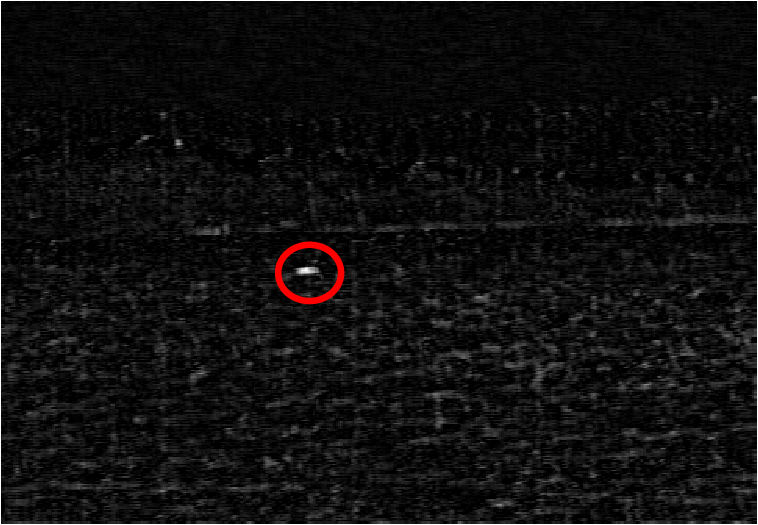}%
   		\label{fig:ldisdv_log}}
	\\
   	\subfloat[]{\includegraphics[width=1.8in]{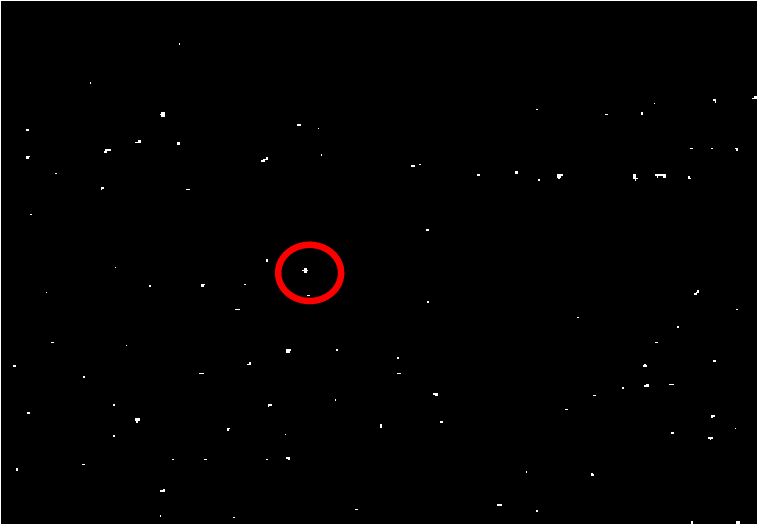}%
   		\label{fig:ldisdv_log_k4}}
   	~
   	\subfloat[]{\includegraphics[width=1.8in]{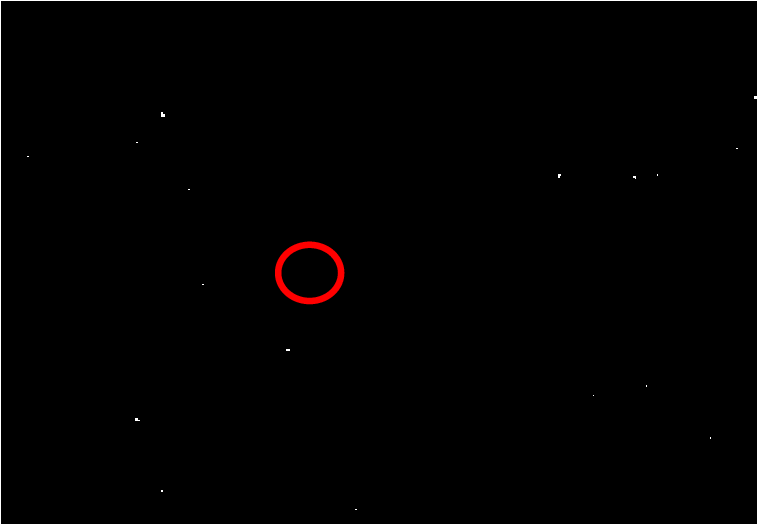}%
   		\label{fig:ldisdv_log_k5}}	
   	\caption{Shortcoming of local thresholding. a) Original input image, the target area is marked by red ellipse, b) the result of target enhancement using multi-scale Laplacian of Gaussian  (LoG) method, c) the local thresholding applied on (b) with $k=4$, c) the local thresholding applied on (b) with $k=5$. }
   	\label{fig:ldisdv_log_dem}
   \end{figure}

\autoref{fig:thresh_Vs_w} shows the local threshold value which is normalized to the target amplitude ($\frac{T}{A}$) versus different control parameter. It is clear that, only when the ($\frac{T}{A}$) fraction is less than one the target can be detected correctly. Another finding which can be derived from the figure is that the maximum control parameter decreases as the target width increases. Therefore, unlike the global case, the effective control parameter to extract real targets and eliminate background clutter depends on the target area in the saliency map. Also, the reasonable rang for control parameter is narrowed when the local neighborhood is decreased (\autoref{fig:thresh_Vs_w17}). Moreover, using \autoref{eq:adaptive_local_thresh} to extract true target from saliency map leads to many false alarms. \autoref{fig:ldisdv_log_dem} shows the   local thresholding on the saliency map of multi-scale  Laplacian of Gaussian  (LoG) method \cite{kim2012scale}. As shown in the \autoref{fig:ldisdv_log}, the target area is the most salient region in saliency map. However, after thresholding using local method (\autoref{fig:ldisdv_log_k4}), there are too many false responses. The only way to limit false responses to an acceptable range is to increase the control parameter. However, the true target is not extracted when the local threshold is increased. Note that there are still too many false responses in     \autoref{fig:ldisdv_log_k5}.

Based on the local thresholding results \autoref{fig:ldisdv_log_dem}, this method (\autoref{eq:adaptive_local_thresh}) is not a proper strategy to discriminate  target area from background clutter. 
%
%
\section{Post-thresholding evaluation}
After applying a predefined threshold to the saliency map, a binary image is obtained. In this case, the  prevalent metrics to evaluate the performance of the detection algorithms are probability of false-alarm $P_{fa}$ and detection $P_d$. These two metrics are defined as \cite{moradi2016scale}:
\begin{equation}
  P_{fa}=\frac{N_f}{N_{tot}}   \quad,\quad  P_d=\frac{N_d}{N_r}
   \label{eq:pfa_pd}
\end{equation}
where
   $N_f$,
   $N_{tot}$, 
   $N_d$,
and   $N_r$
 denote the number of wrongly detected pixels, the total number of pixels, the number of pixels which are detected correctly, and the target pixels in the ground-truth, respectively. The receiver operational characteristics   (ROC) curve is constructed by considering each    $(P_{fa},P_d)$ pair at different threshold level.

    \begin{figure}[!t]
   	\centering
   	\includegraphics[width=1.6in]{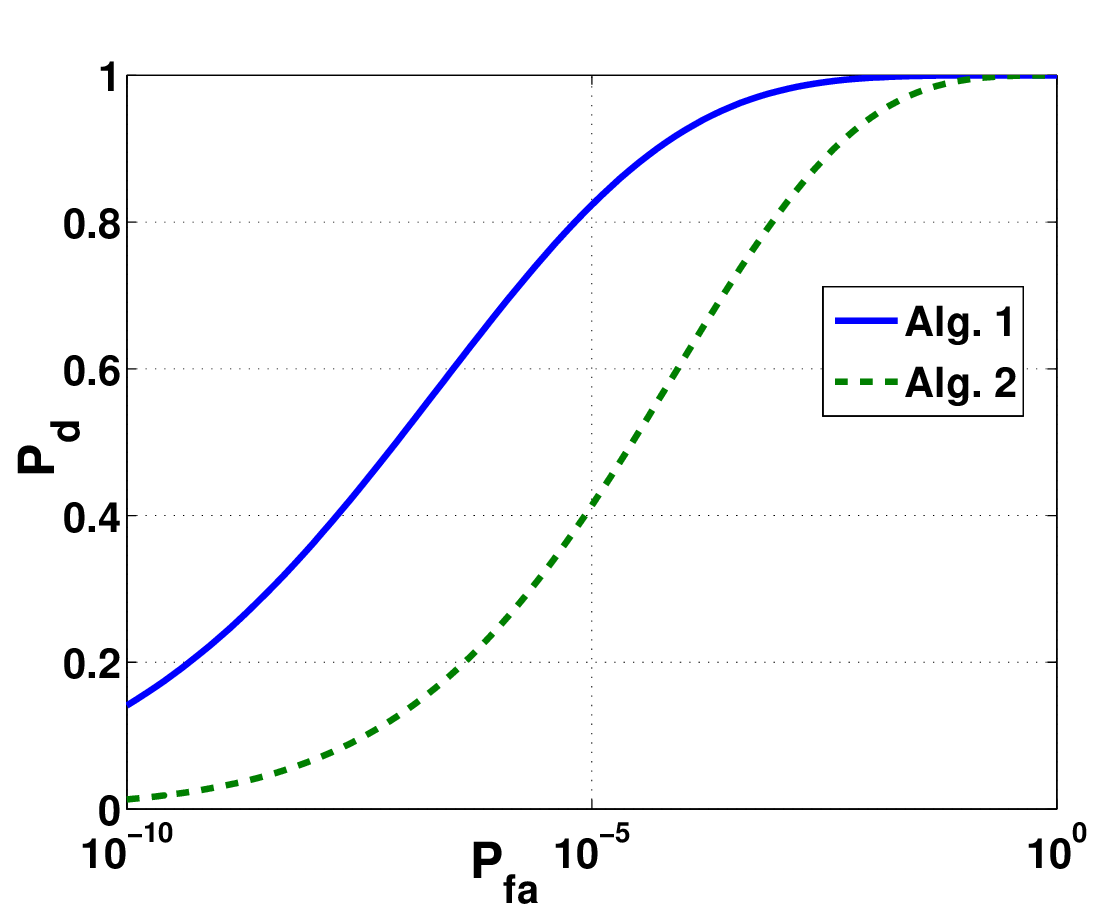}%
   	   	\caption{  ROC curve for two typical detectors}
   	\label{fig:ROC_exm_eval}
   \end{figure}

\begin{figure}[t!]
   	\centering
   	\subfloat[]{\includegraphics[width=1.4in]{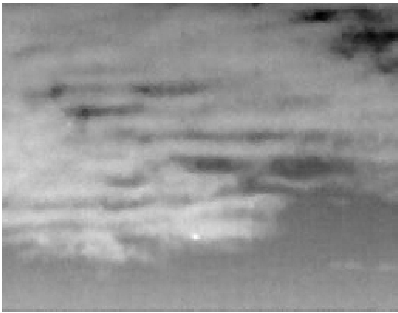}%
   		\label{fig:input_img_oldroc}}
   	~~
   	\subfloat[]{\includegraphics[width=1.4in]{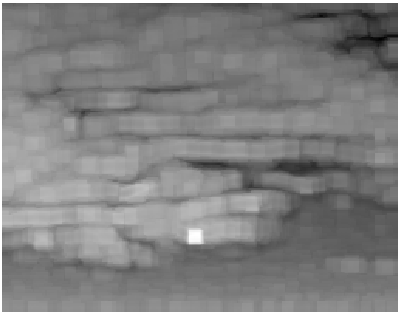}%
   		\label{fig:lcm_img_oldroc}}
   	~~
   	\subfloat[]{\includegraphics[width=1.4in]{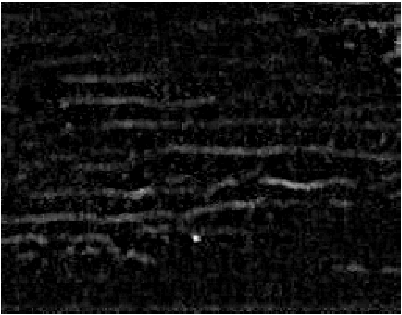}%
   		\label{fig:tophat_img_oldroc}}
   	\caption{a) original image, b) the LCM filtering result, c) the Top-Hat filtering result. }
   	\label{fig:tophat_vs_lcm_filter}
   \end{figure}

\begin{figure}[t!]
   	\centering
   	\subfloat[]{\includegraphics[width=2.2in]{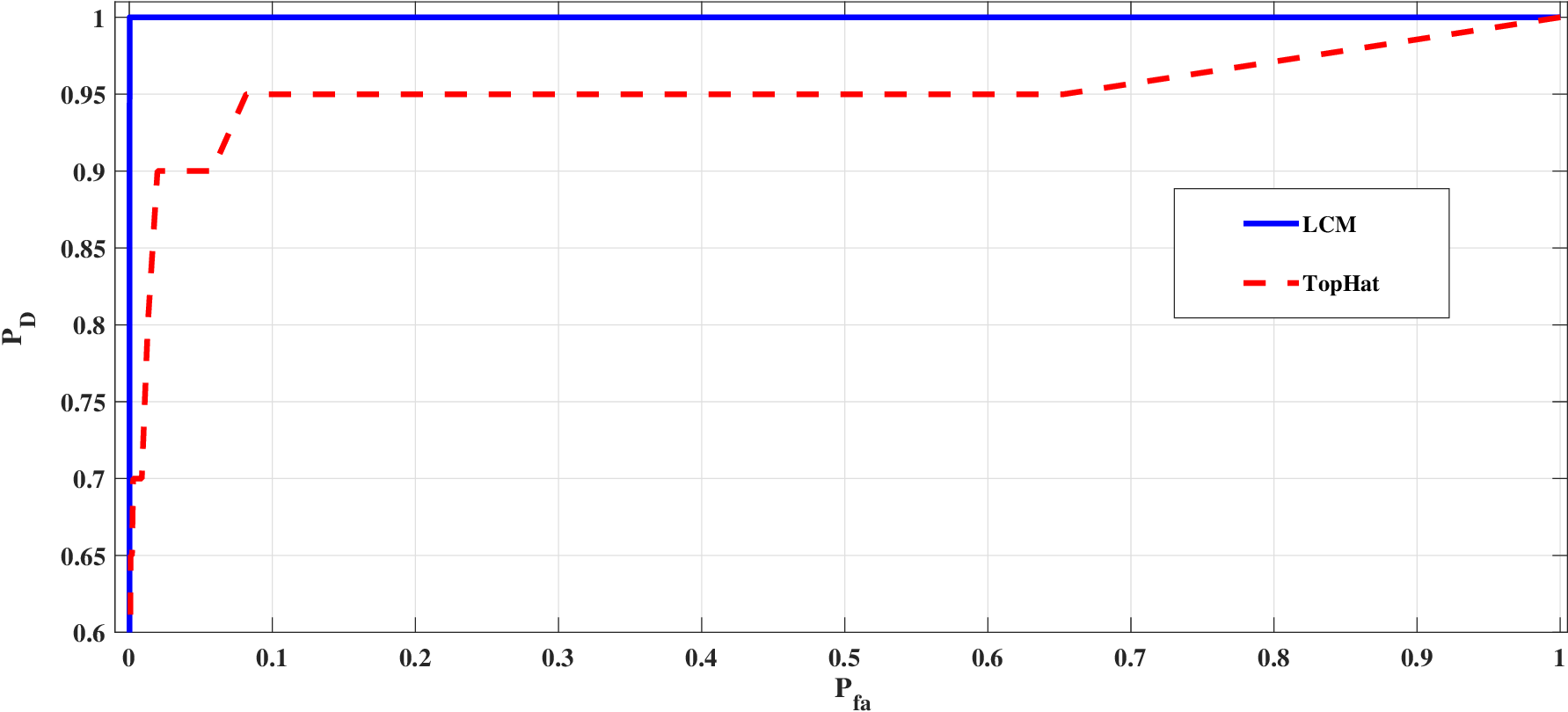}%
   		\label{fig:lcm_vs_tophat_oldroc}}
~
   	\subfloat[]{\includegraphics[width=2.2in]{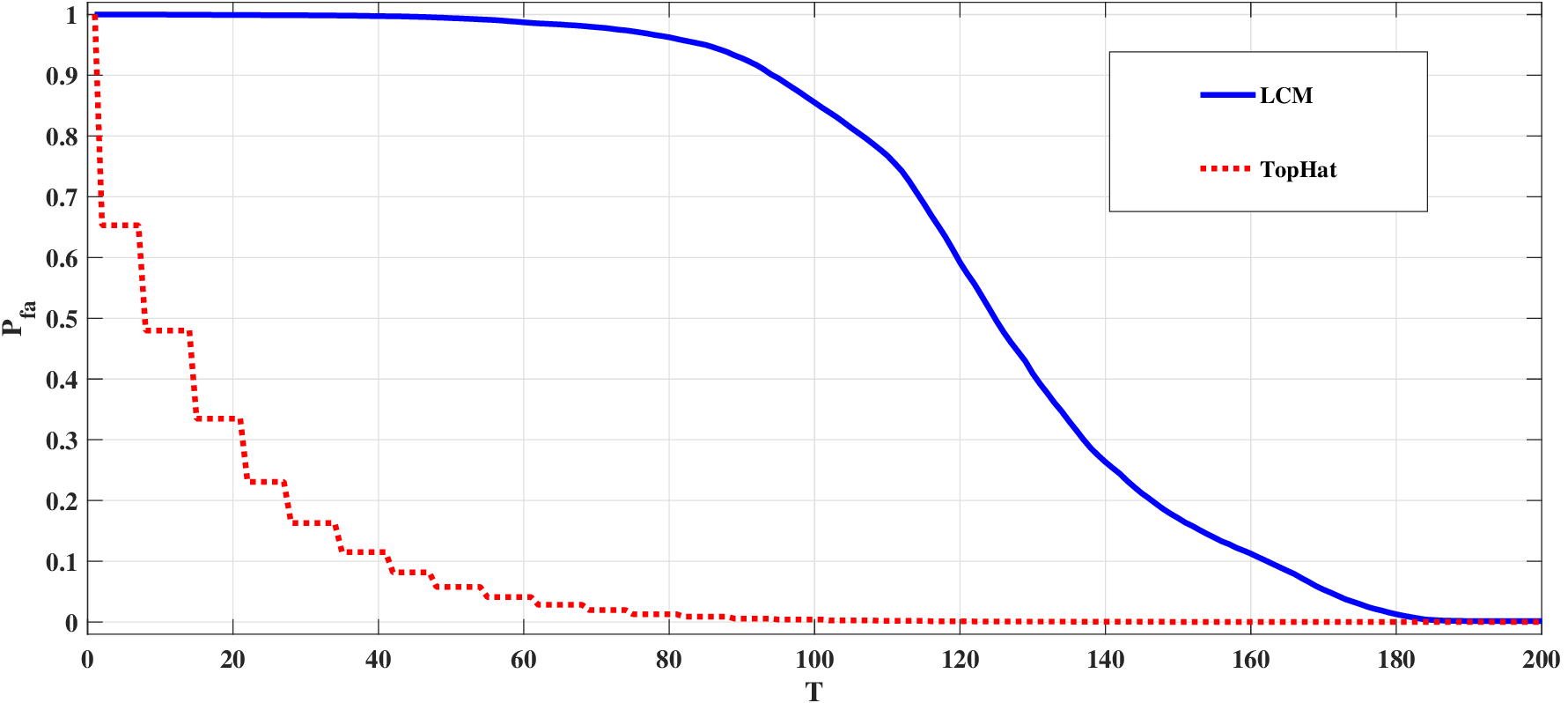}%
   		\label{fig:lcm_vs_tophat_pfa}}
   	\caption{a) The ROC curve, b) false-alarm rate versus different threshold levels.  }
   	\label{fig:tophat_vs_lcm_roc_pfa}
   \end{figure}

\autoref{fig:ROC_exm_eval} shows the  ROC curve for two typical detectors. As shown in the figure, for a constant false-alarm rate, the detector $\#1$ has higher detection rate, and outperforms the  algorithm $\#2$. The ROC curve is a satisfactory tool to evaluate the  performance of different detectors. However,  if the detection rate and false-alarm rate are not defined accurately, the final ROC curve is not   a reliable measure anymore.  In order to demonstrate the deficiency of the definitions of the $P_d$ and $P_{fa}$ (\autoref{eq:pfa_pd}), let consider the target detection ability of two well-known small infrared target detection  algorithms; Local contrast method (LCM) \cite{chen2013local} and Top-hat algorithm \cite{tom1993morphology}. \autoref{fig:tophat_vs_lcm_filter}  shows the detection results of these two algorithms. As shown in the figure, the Top-hat filtering method clearly outperforms the LCM algorithm. however, the  ROC curve gives contradictory result against visual perception (\autoref{fig:lcm_vs_tophat_oldroc}). Also, by constructing  the curve of the false-alarm rate versus different threshold levels (\autoref{fig:lcm_vs_tophat_pfa}), the low performance of the LCM algorithm is clearly seen.  Therefore, the former definition of the  $P_d$  (\autoref{eq:pfa_pd}) is not appropriate for this crucial metric.
 
\begin{figure}[t!]
   	\centering
   	\subfloat[]{\includegraphics[width=2.5in]{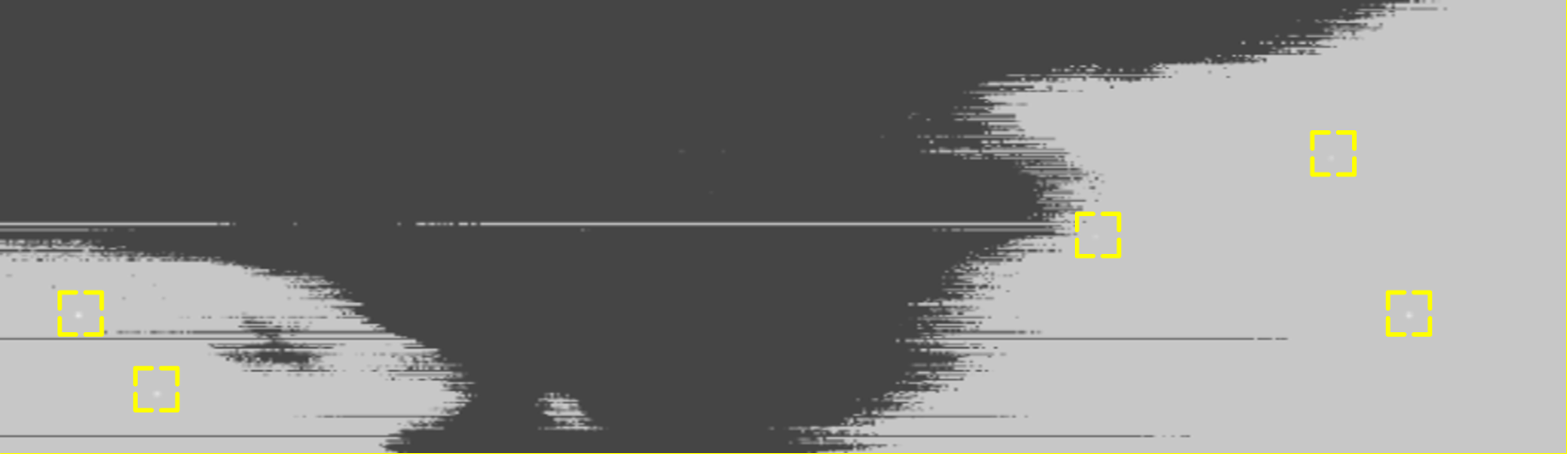}%
   		\label{fig:charachter_input1}}
~
   	\subfloat[]{\includegraphics[width=2.5in]{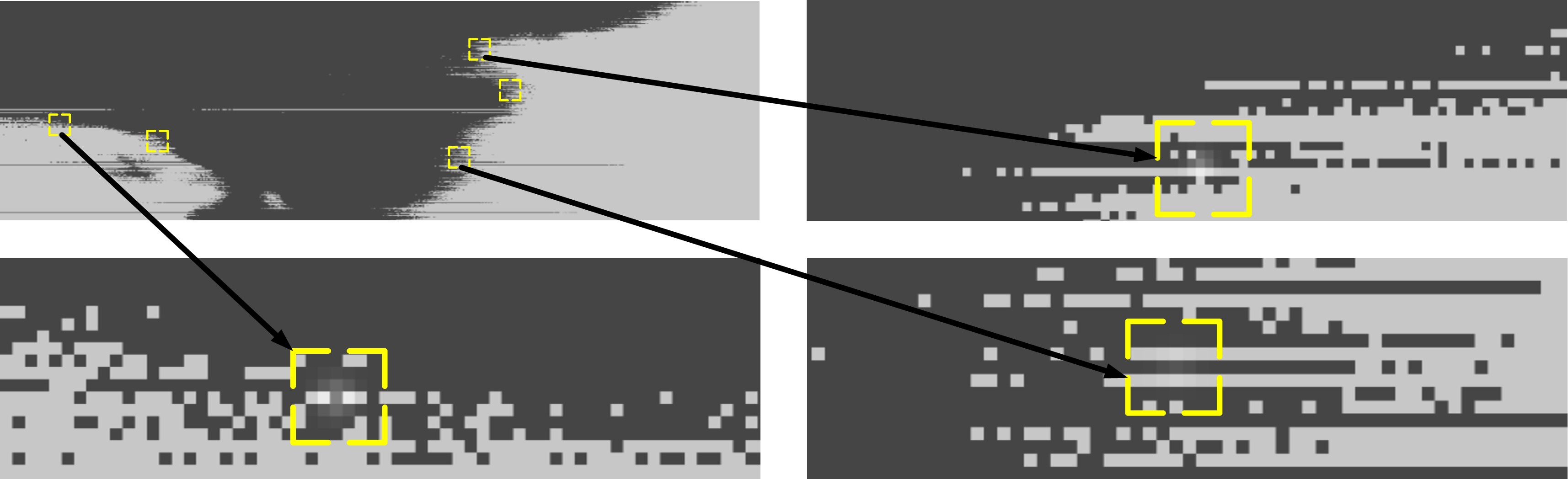}%
   		\label{fig:charachter_input2}}
   		\\
   		   	\subfloat[]{\includegraphics[width=2.5in]{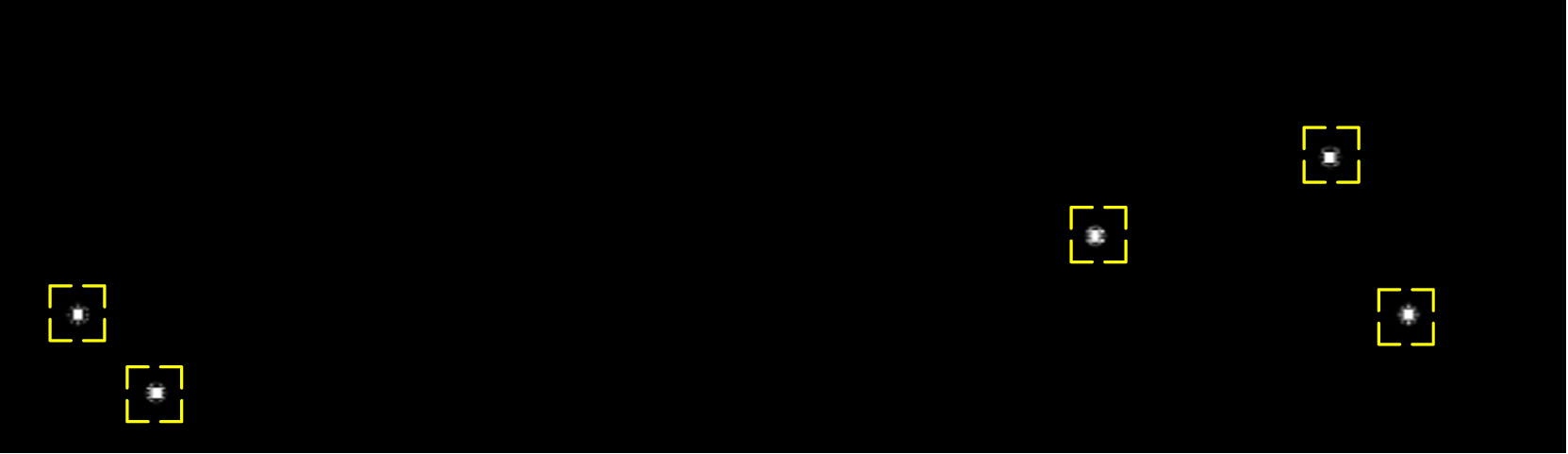}%
   		\label{fig:charachter_output1}}
~
   	\subfloat[]{\includegraphics[width=2.5in]{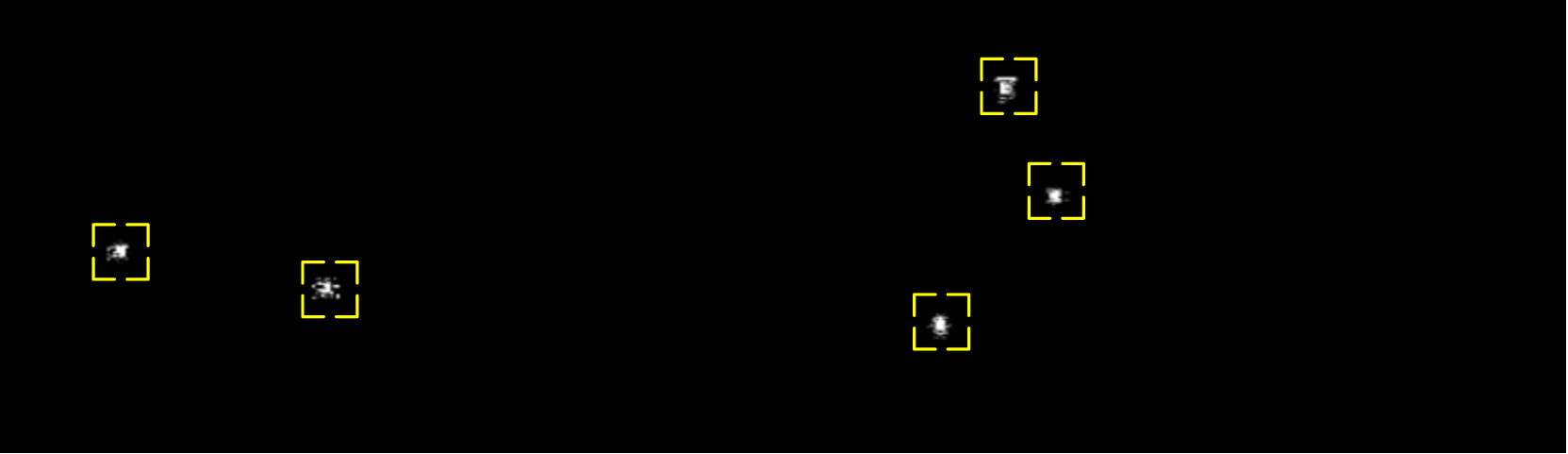}%
   		\label{fig:charachter_output2}}
   	\caption{a) Synthetic targets in homogeneous local background, b) low contrast  targets  in background clutter edges, c) the character filter response to \subref{fig:charachter_input1} and \subref{fig:charachter_input2}). }
   	\label{fig:charachter_input_putput}
   \end{figure}

\begin{figure}[t!]
   	\centering
   	\subfloat[]{\includegraphics[width=1.4in]{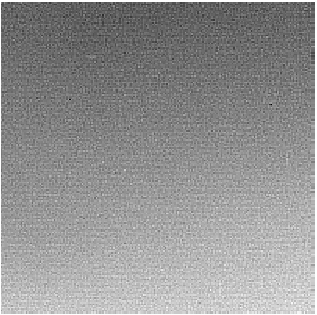}%
   		\label{fig:charachter_real_input1}}
   	~
   	\subfloat[]{\includegraphics[width=1.4in]{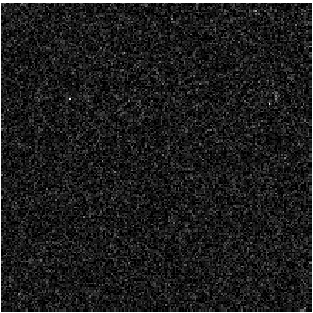}%
   		\label{fig:charachter_real_output1}}
   	\\
   	\subfloat[]{\includegraphics[width=1.4in]{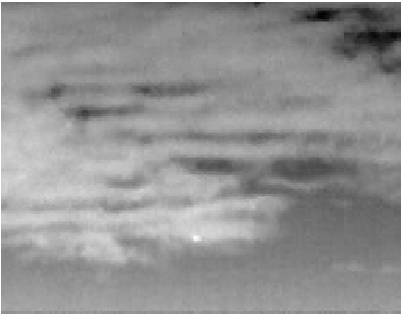}%
   		\label{fig:charachter_real_input2}}
   	~
   	\subfloat[]{\includegraphics[width=1.4in]{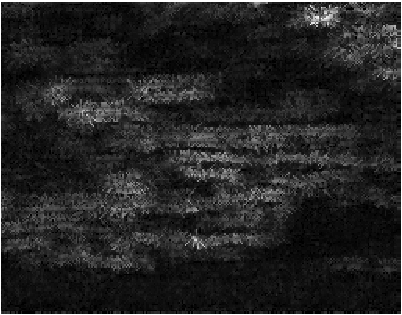}%
   		\label{fig:charachter_real_output2}}
   	\caption{a, c) real infrared scenario, b , d) the response of character filter \cite{hu2011infrared} to \subref{fig:charachter_real_input1} and \subref{fig:charachter_real_input2}, respectively. }
   	\label{fig:charachter_real_output}
   \end{figure}

Another alternative definition for $P_d$ is suggested in the literature (\cite{han2018infrared}):
   \begin{equation}
   P_d=\frac{N_D}{N_R}
   \end{equation}
where   $N_D$
and   $N_R$ are number of detected true targets, and total number of true targets. While this new definition addresses the deficiency of the former one (\autoref{eq:pfa_pd}), there are still some drawbacks regrading this formula; The real infrared scenarios usually contain limited number of targets. To overcome this drawback, synthetic targets are usually created using Gaussian spatial distribution.  However, spatial distribution-based target detection algorithms directly benefit from synthetic data, so the final evaluation is not fair. An example is provided here to better demonstration of this situation. The character filter \cite{hu2011infrared} utilizes Gaussian spatial distribution as a measure to distinguish  between real target and background clutter. As shown in \autoref{fig:charachter_input_putput}, when the small target has exactly  Gaussian distribution, the character filter effectively can enhance the small targets and eliminate background clutter. However, in real infrared scenarios, which the spatial distribution of small targets does not follow the Gaussian distribution \cite{moradi2016scale}, the detection results of character filter is chaotic (\autoref{fig:charachter_real_output}).

According to aforementioned   issues regarding the post-thresholding performance evaluation metrics, herein, awe present a new approach capable of addressing all the shortcomings. Since in a successful detection operation, at least one pixel is detected after thresholding operation, the following procedure is introduced to obtain new post-thresholding performance measure:
\begin{enumerate}[label=\roman*)]
\item The upper bound for control parameter  ($k_{\max}$) is calculated (\autoref{eq:rec_scr_new_kup}). 
\item  The $\lbrack 0-k_{max}\rbrack$ interval is chosen as valid interval for performance evaluation.
\item For each different control parameters, the false-alarm rate is calculated using  \autoref{eq:pfa_pd}.
\item The  false-alarm rate versus control parameter ($P_{fa}$ -- $k$) curve is constructed. In the next step, the $[0$ -- $k]$ interval is linearly mapped to $[0$ -- $1]$ range. This normalization allows us to fairly compare and evaluate different algorithms.
\end{enumerate}

After constructing ($P_{fa}$ -- $k$) curve, the following measures can be extracted:
\begin{itemize}
\item The maximum control parameter ($k_{\max}$) is the first inferred performance evaluation metric. The larger $k_{\max}$, the higher detection ability.
\item The false-alarm rate at $k_{\max}$, which is called $P_{fa,\min}$ here,  is the second evaluation metric. It is obvious that the false-alarm rate of the system can not be less than $P_{fa,\min}$ while the true target is detected.
\end{itemize}

After normalizing $[0$ -- $k]$ interval  to $[0$ -- $1]$ range, the false alarm rate of the detection algorithms can be plotted in single figure. Then, the algorithm with satisfying detection performance can be chosen for the practical application.
\section{Detection ability evaluation using new metrics}

\begin{figure*}[h!]
	\centering
	\subfloat{\includegraphics[width=1.1in]{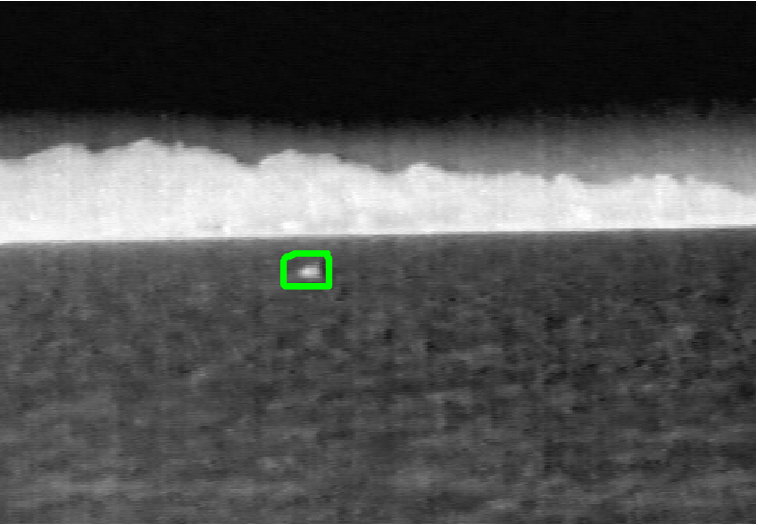}%
		\label{fig:det_res_orig1}}
	~
	\subfloat{\includegraphics[width=1.1in]{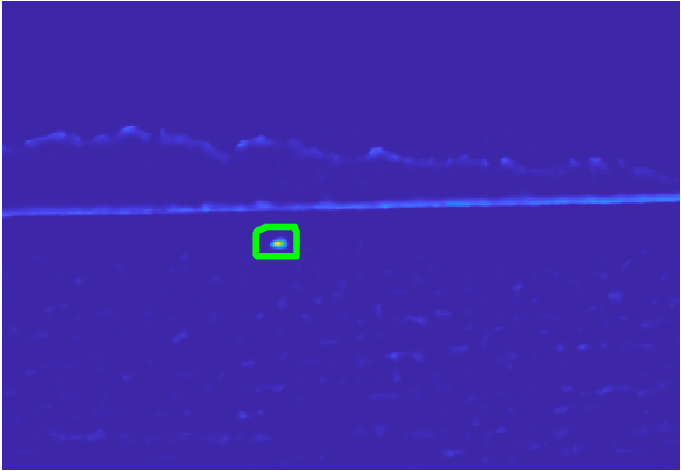}%
		\label{fig:det_res_aagd1}}
	~
	\subfloat{\includegraphics[width=1.1in]{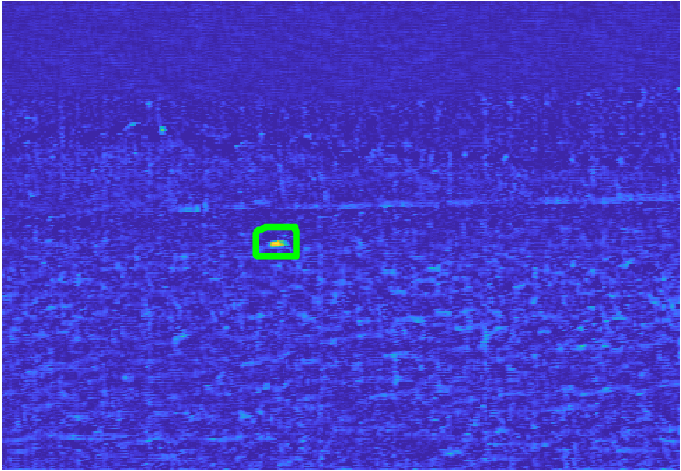}%
		\label{fig:det_res_tophat1}}
	~
		\subfloat{\includegraphics[width=1.1in]{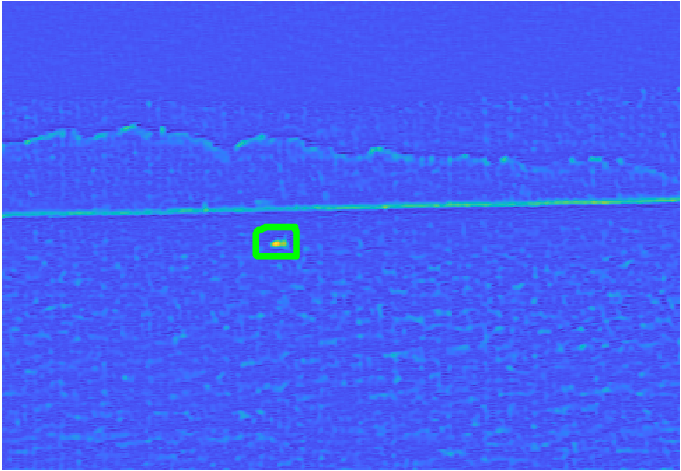}%
		\label{fig:det_res_log1}}
	~
	\subfloat{\includegraphics[width=1.1in]{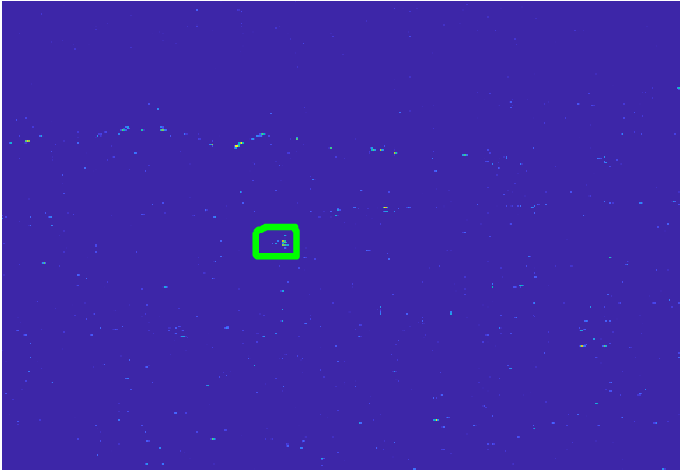}%
		\label{fig:det_res_NIPPS1}}
			~
	\subfloat{\includegraphics[width=1.1in]{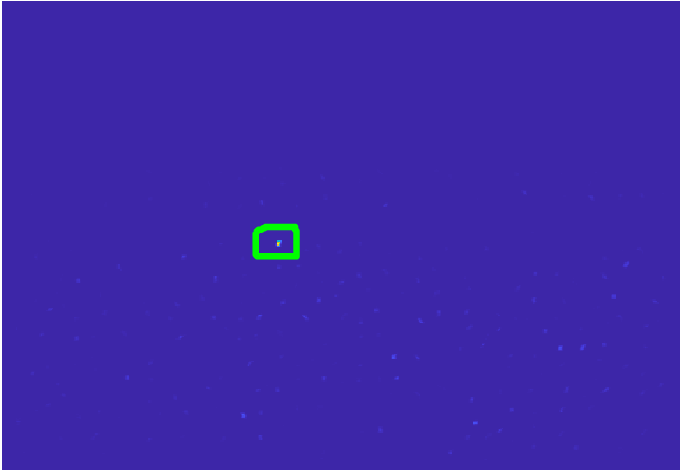}%
		\label{fig:det_res_admd1}}
	\\
		\subfloat{\includegraphics[width=1.1in]{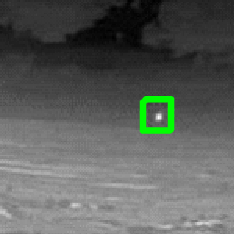}%
		\label{fig:det_res_orig2}}
	~
	\subfloat{\includegraphics[width=1.1in]{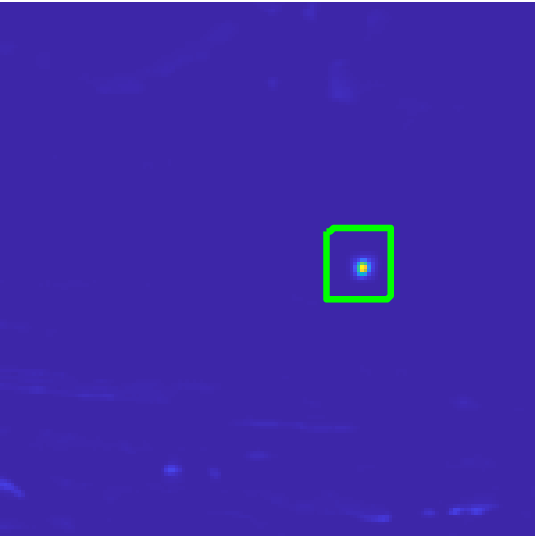}%
		\label{fig:det_res_aagd2}}
	~
	\subfloat{\includegraphics[width=1.1in]{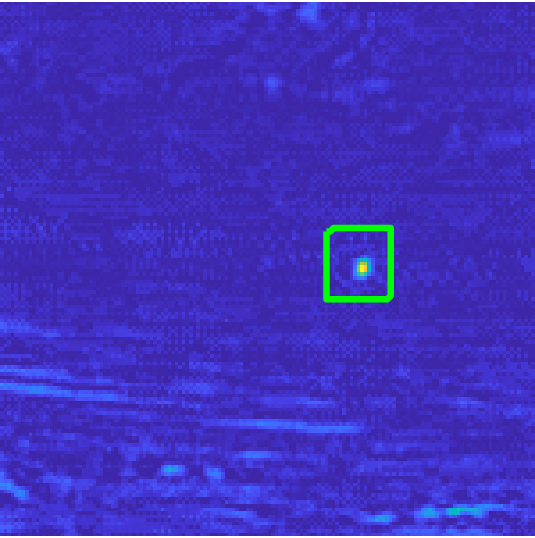}%
		\label{fig:det_res_tophat2}}
	~
		\subfloat{\includegraphics[width=1.1in]{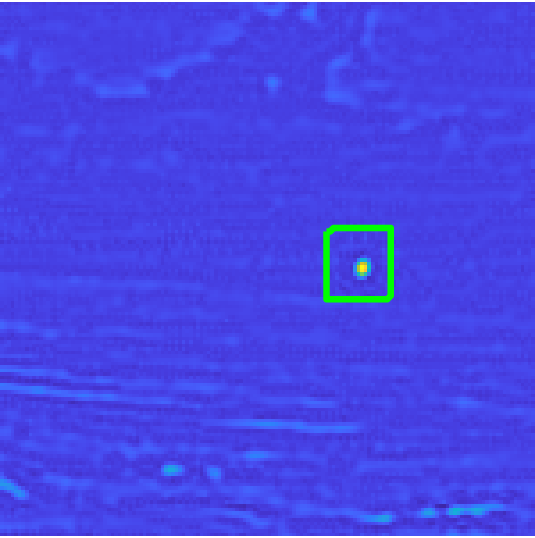}%
		\label{fig:det_res_log2}}
	~
	\subfloat{\includegraphics[width=1.1in]{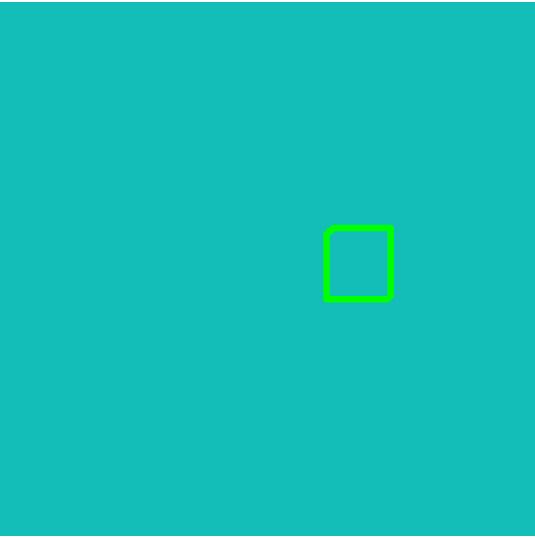}%
		\label{fig:det_res_NIPPS2}}
			~
	\subfloat{\includegraphics[width=1.1in]{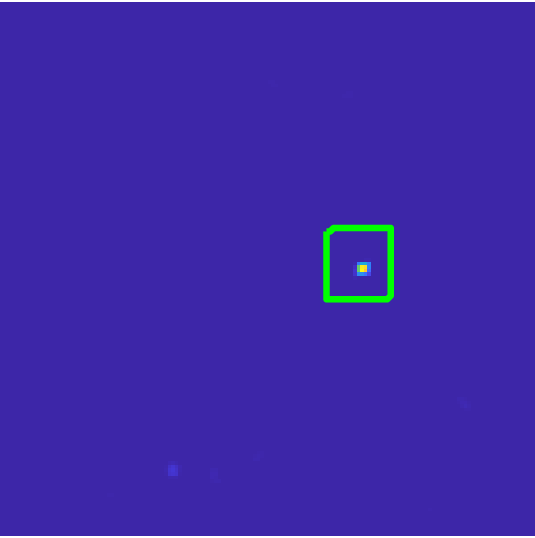}%
		\label{fig:det_res_admd2}}
		\\
			\subfloat{\includegraphics[width=1.1in]{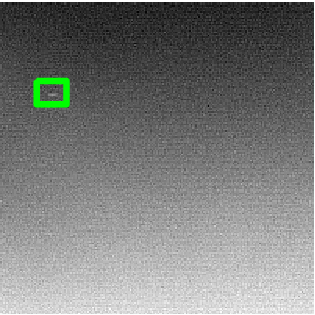}%
		\label{fig:det_res_orig3}}
	~
	\subfloat{\includegraphics[width=1.1in]{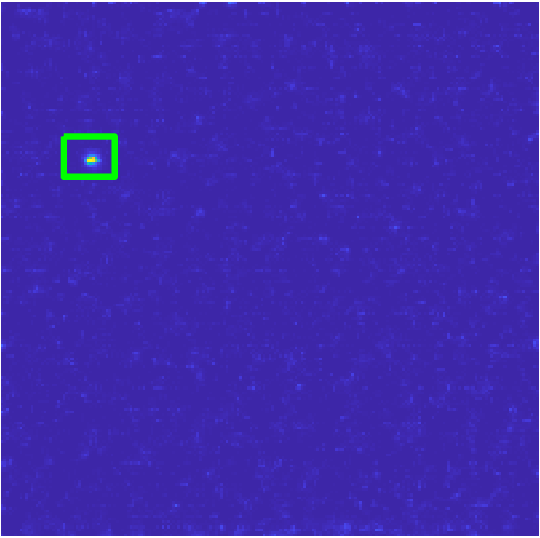}%
		\label{fig:det_res_aagd3}}
	~
	\subfloat{\includegraphics[width=1.1in]{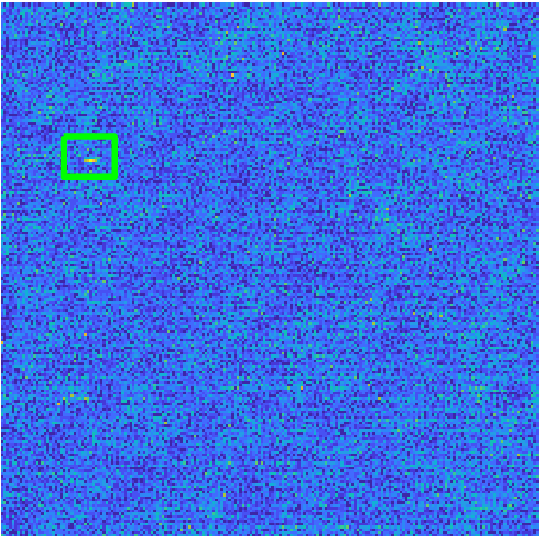}%
		\label{fig:det_res_tophat3}}
	~
		\subfloat{\includegraphics[width=1.1in]{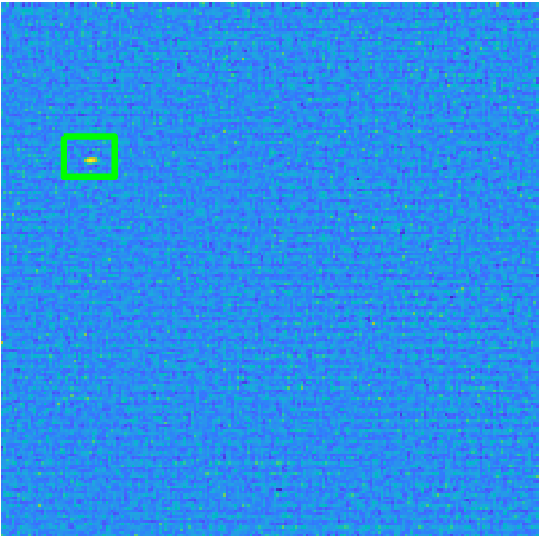}%
		\label{fig:det_res_log3}}
	~
	\subfloat{\includegraphics[width=1.1in]{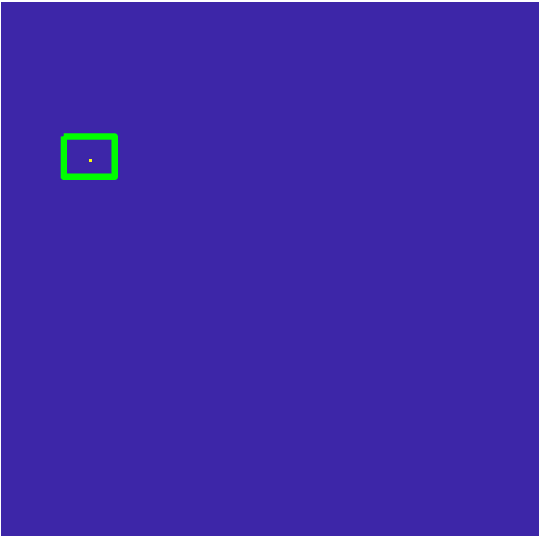}%
		\label{fig:det_res_NIPPS3}}
			~
	\subfloat{\includegraphics[width=1.1in]{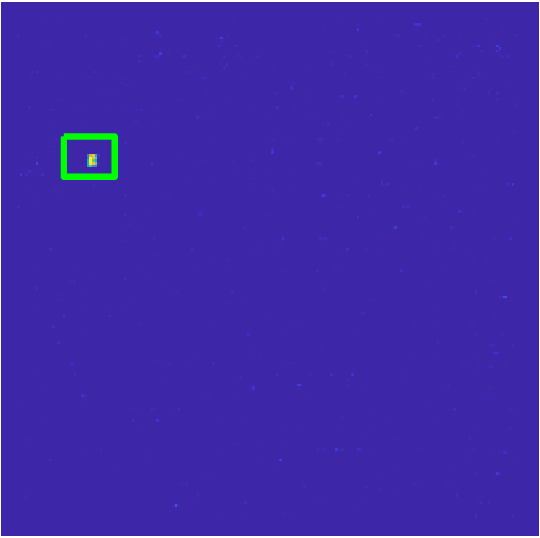}%
		\label{fig:det_res_admd3}}
				\\
			\subfloat{\includegraphics[width=1.1in]{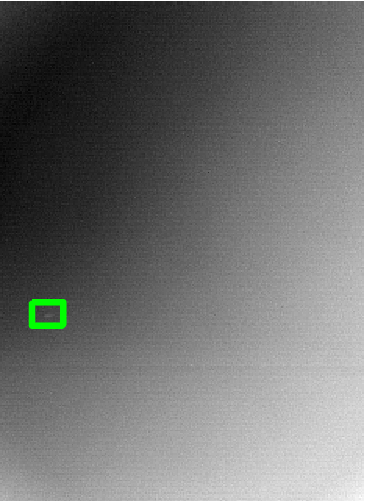}%
		\label{fig:det_res_orig4}}
	~
	\subfloat{\includegraphics[width=1.1in]{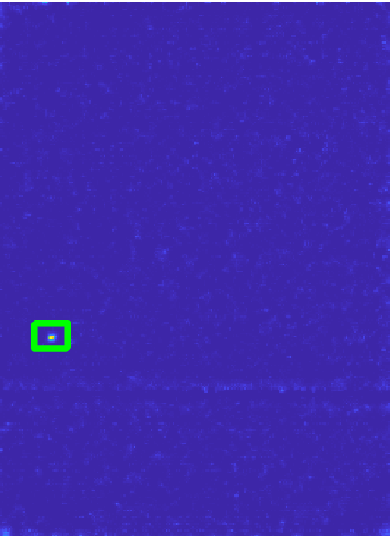}%
		\label{fig:det_res_aagd4}}
	~
	\subfloat{\includegraphics[width=1.1in]{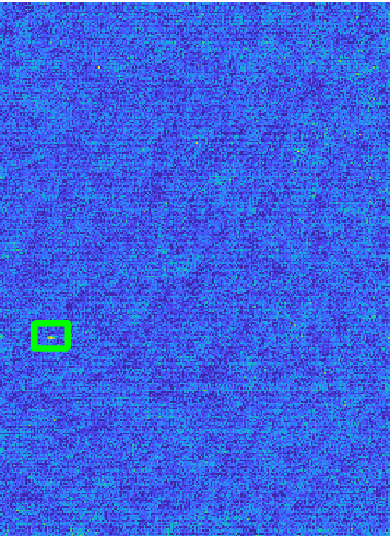}%
		\label{fig:det_res_tophat4}}
	~
		\subfloat{\includegraphics[width=1.1in]{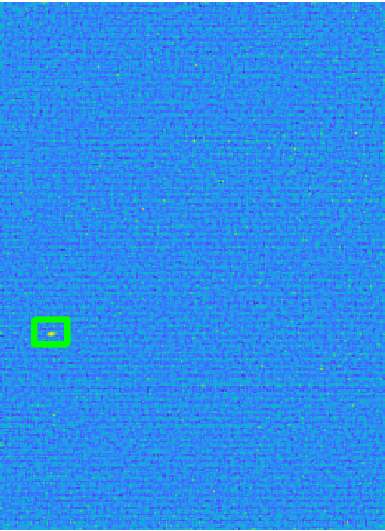}%
		\label{fig:det_res_log4}}
	~
	\subfloat{\includegraphics[width=1.1in]{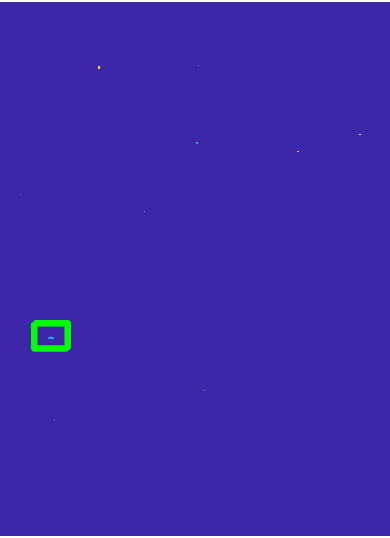}%
		\label{fig:det_res_NIPPS4}}
			~
	\subfloat{\includegraphics[width=1.1in]{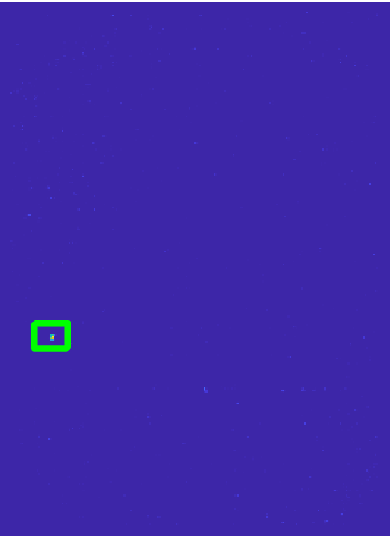}%
		\label{fig:det_res_admd4}}
				\\
			\subfloat{\includegraphics[width=1.1in]{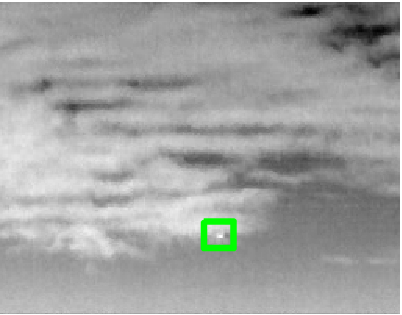}%
		\label{fig:det_res_orig5}}
	~
	\subfloat{\includegraphics[width=1.1in]{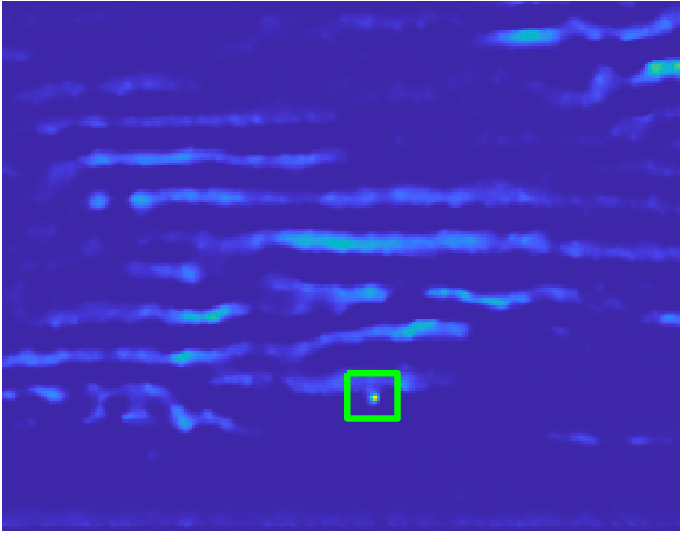}%
		\label{fig:det_res_aagd5}}
	~
	\subfloat{\includegraphics[width=1.1in]{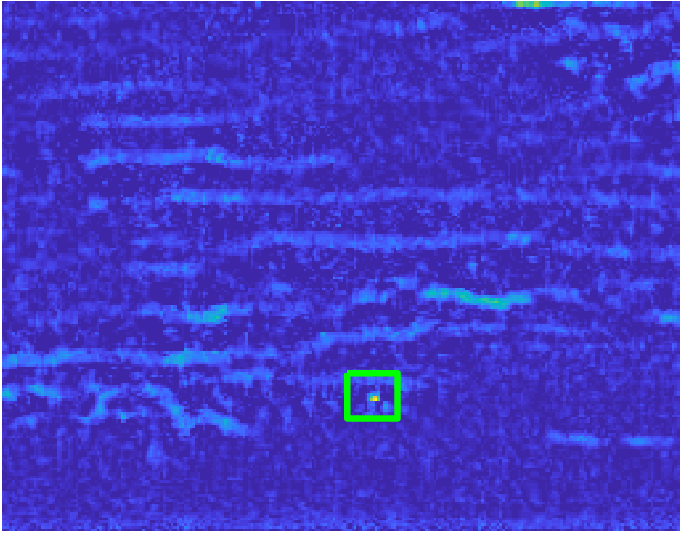}%
		\label{fig:det_res_tophat5}}
	~
		\subfloat{\includegraphics[width=1.1in]{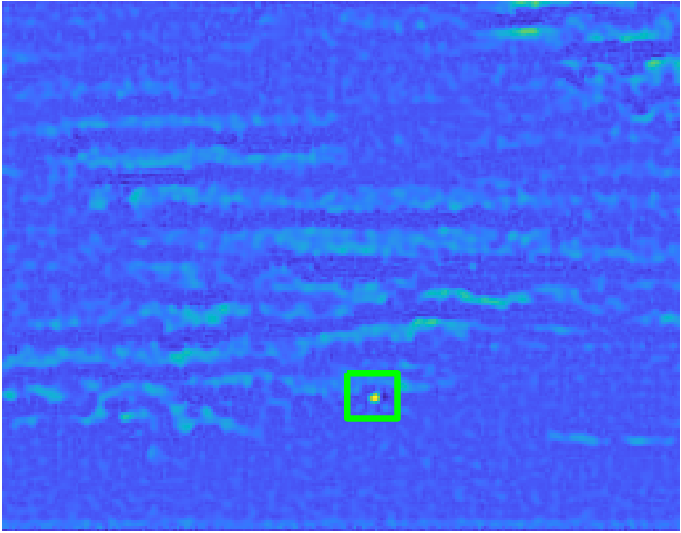}%
		\label{fig:det_res_log5}}
	~
	\subfloat{\includegraphics[width=1.1in]{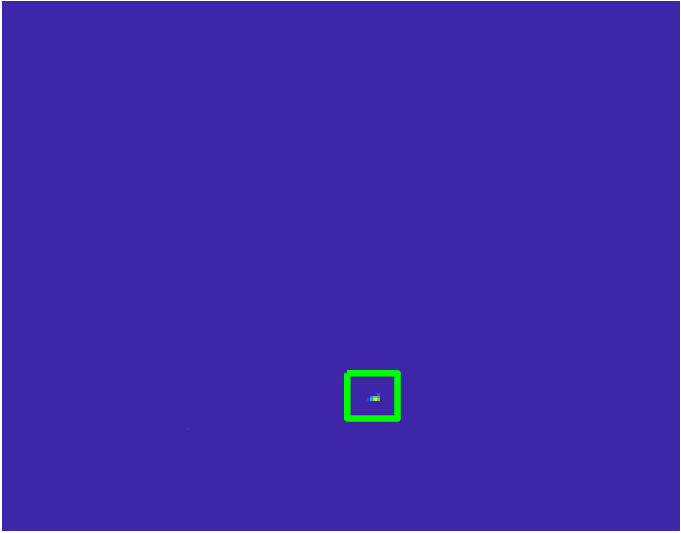}%
		\label{fig:det_res_NIPPS5}}
			~
	\subfloat{\includegraphics[width=1.1in]{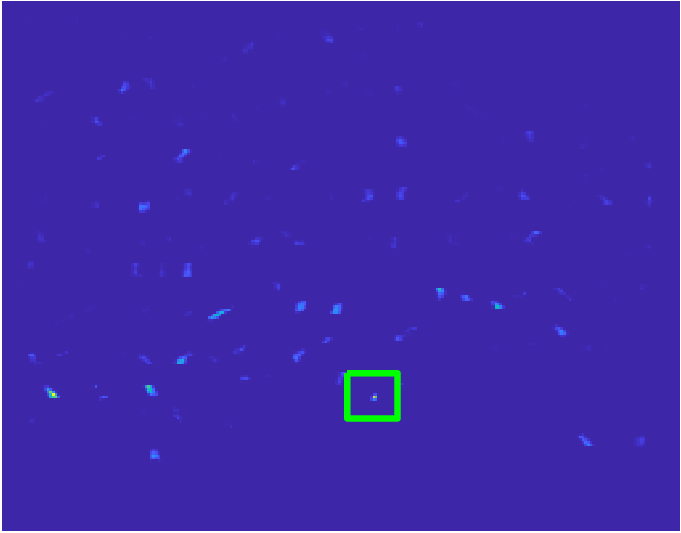}%
		\label{fig:det_res_admd5}}
				\\
			\subfloat{\includegraphics[width=1.1in]{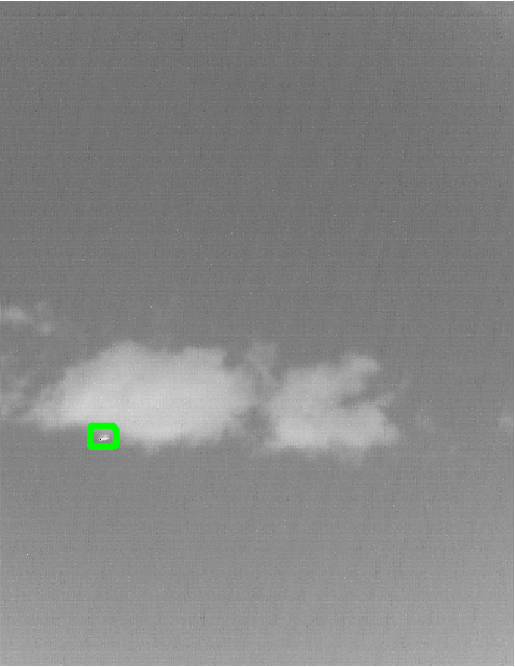}%
		\label{fig:det_res_orig6}}
	~
	\subfloat{\includegraphics[width=1.1in]{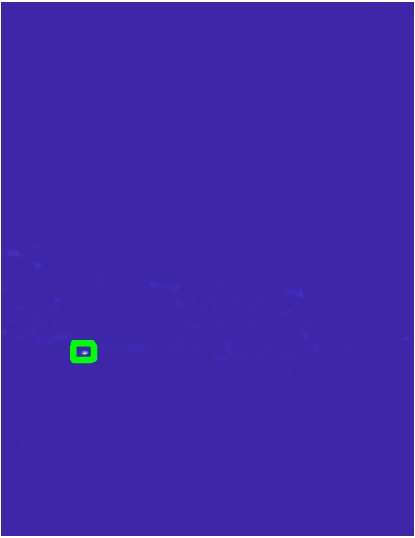}%
		\label{fig:det_res_aagd6}}
	~
	\subfloat{\includegraphics[width=1.1in]{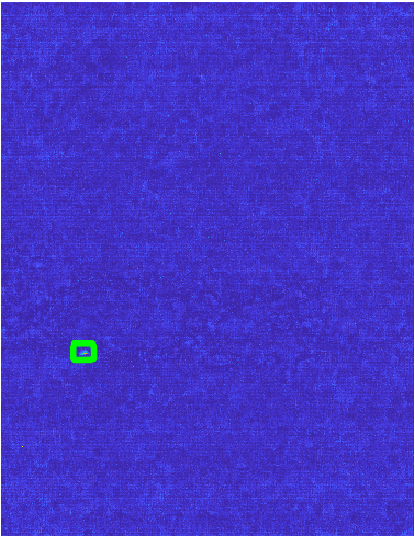}%
		\label{fig:det_res_tophat6}}
	~
		\subfloat{\includegraphics[width=1.1in]{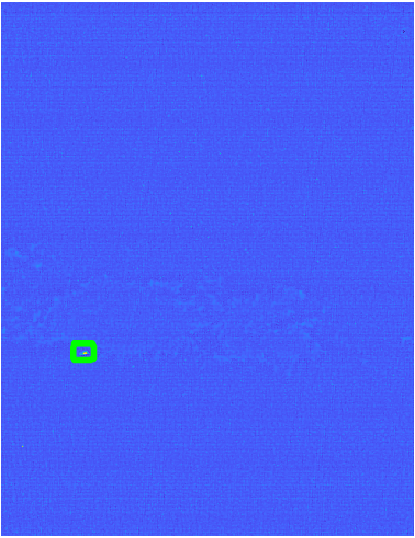}%
		\label{fig:det_res_log6}}
	~
	\subfloat{\includegraphics[width=1.1in]{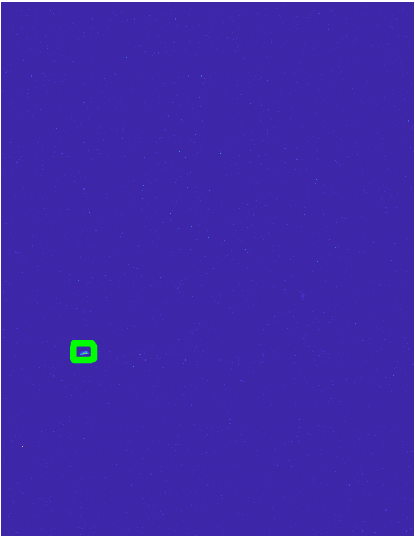}%
		\label{fig:det_res_NIPPS6}}
			~
	\subfloat{\includegraphics[width=1.1in]{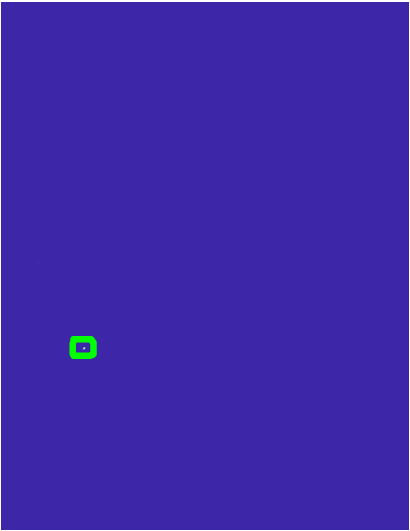}%
		\label{fig:det_res_admd6}}
	\caption{Pre-thresholding results of the algorithms under the test on real infrared images (Target region is marked by yellow circle). From the left: { \bf the first column:} original images, {\bf the second column:} filtering results of AAGD algorithm,  {\bf the third column:} filtering results of Tophat transform, {\bf the fourth column:} filtering results of LoG algorithm, {\bf the fifth column:} filtering results of NIPPS algorithm, and {\bf the sixth column:} filtering results of ADMD algorithm.}
	\label{fig:real_test}
\end{figure*}

In order to evaluate the detection ability using the proposed metrics, five well-known small infrared target detection algorithms are chosen to conduct the experiments. \autoref{tab:base_alg} reports the baseline algorithms and their implementation details. The pre-thresholding enhancement results of each algorithm are depicted in \autoref{fig:real_test}. Visually speaking, the ADMD and NIPPS algorithms have better performance in background suppression (the background region is mapped to zero value). NIPPS algorithm fails to detect the target in the seconf test image. Also, ADMD shows poor performance in detecting target close to clutter edge (\autoref{fig:real_test}). AAGD shows a moderate performance in terms of background suppression and noise elimination since this algorithm benefits from local averaging.   LoG and TopHat algorithms show poor performance in this regard. 
 LoG and TopHat filters are sensitive to noise and sharp edges, therefore, there are  too many false responses in their saliency maps.

The results of evaluation using new metrics are reported in \autoref{tab:kmax} and \autoref{tab:pfamin}. As reported in \autoref{tab:kmax}, again, ADMD and NIPPS algorithms have better enhancement for target area. However, The minimum false alarm  $P_{fa,\min}$ is not zero for one case for AMDD and in four cases for NIPPS. Therefore, the ADMD algorithm outperforms other baselines.

\begin{table}
\centering
\caption{The baseline algorithms}
\resizebox{0.95\linewidth}{!}{%
\begin{tabular}{ll} 
\hline
\textbf{Detection Algorithm} & \textbf{Details}               \\ 
\hline
Top-Hat    \cite{tom1993morphology}         & $7\times 7$ structural element  \\
LoG    \cite{kim2012scale}             & With {[}0.50, 0.60, 0.72, 0.86, 1.03, 1.24, 1.49, 1.79, 2.14, 2.57, 3.09, 3.71{]} scale parameters                    \\
AAGD    \cite{deng2016infrared}            &With {[}$3\times 3$, $5\times 5$, $7\times7$, and $9\times 9${]} cell-sizes           \\
NIPPS \cite{dai2017non} & Patch size: $50\times 50$, sliding step: $10$, $L=2$, $r=2\times 10^{-3}$\\
ADMD    \cite{moradi2020fast}             & With {[}$3\times 3$ and $5\times 5${]} cell-sizes         \\
\hline
\end{tabular}}
\label{tab:base_alg}
\end{table}

\begin{table}
\centering
\caption{The value of maximum control parameter $k_{max}$ for different algorithms}
\resizebox{0.95\linewidth}{!}{%
\begin{tabular}{lcccccc} 
\hline
       & the $1^{st}$ test image   & the $2^{nd}$ test image    & the $3^{rd}$ test image     & the $4^{th}$ test image    & the $5^{th}$ test image     & the $6^{th}$ test image        \\ 
\hline
\textbf{AAGD}   & 44.53 & 62.48 & 61.55 & 72.27 & 16.74 & 172.81  \\
\textbf{LoG}    & 16.53 & 28.89 & 8.62  & 8.54  & 13.45 & 37.40   \\
\textbf{TopHat} & 15.88 & 22.89 & 5.64  & 5.87  & 14.44 & 26.91   \\
\textbf{NIPPS}    & 45.23 & NaN & 200.49 & 63.66 & 122.87 & 126.16   \\
\textbf{ADMD}    & 172.94 & 65.85 & 68.55 & 144.39 & 49.98 & 290.93   \\

\hline
\end{tabular}
}
\label{tab:kmax}
\end{table}

\begin{table}
\centering
\caption{The value of minimum probability of false alarm  $ P_{fa,\min}$ for different algorithms}
\resizebox{0.95\linewidth}{!}{%
\begin{tabular}{lcccccc} 
\hline
       & $1^{st}$ test image   & $2^{nd}$ test image    & $3^{rd}$ test image     & $4^{th}$ test image    & $5^{th}$ test image     & $6^{th}$ test image        \\ 
\hline
\textbf{AAGD}   & 0 & 0 & 0 & 0 & 0 & 0  \\
\textbf{LoG}    & 0 & 0 & 0  & 0  &0 & 0   \\
\textbf{TopHat} & 0 & 0& 0  & 2.68e-05 & 0 & 0   \\
\textbf{NIPPS}    & 4.93e-05 & NaN & 6.7e-05 & 1.3412e-5 & 0 & 0   \\
\textbf{ADMD}    & 0 & 0 & 0 & 0 & 1.95e-05 & 0   \\
\hline
\end{tabular}
}
\label{tab:pfamin}
\end{table}

\begin{figure}[t!]
   	\centering
   	\subfloat[]{\includegraphics[width=3.3in]{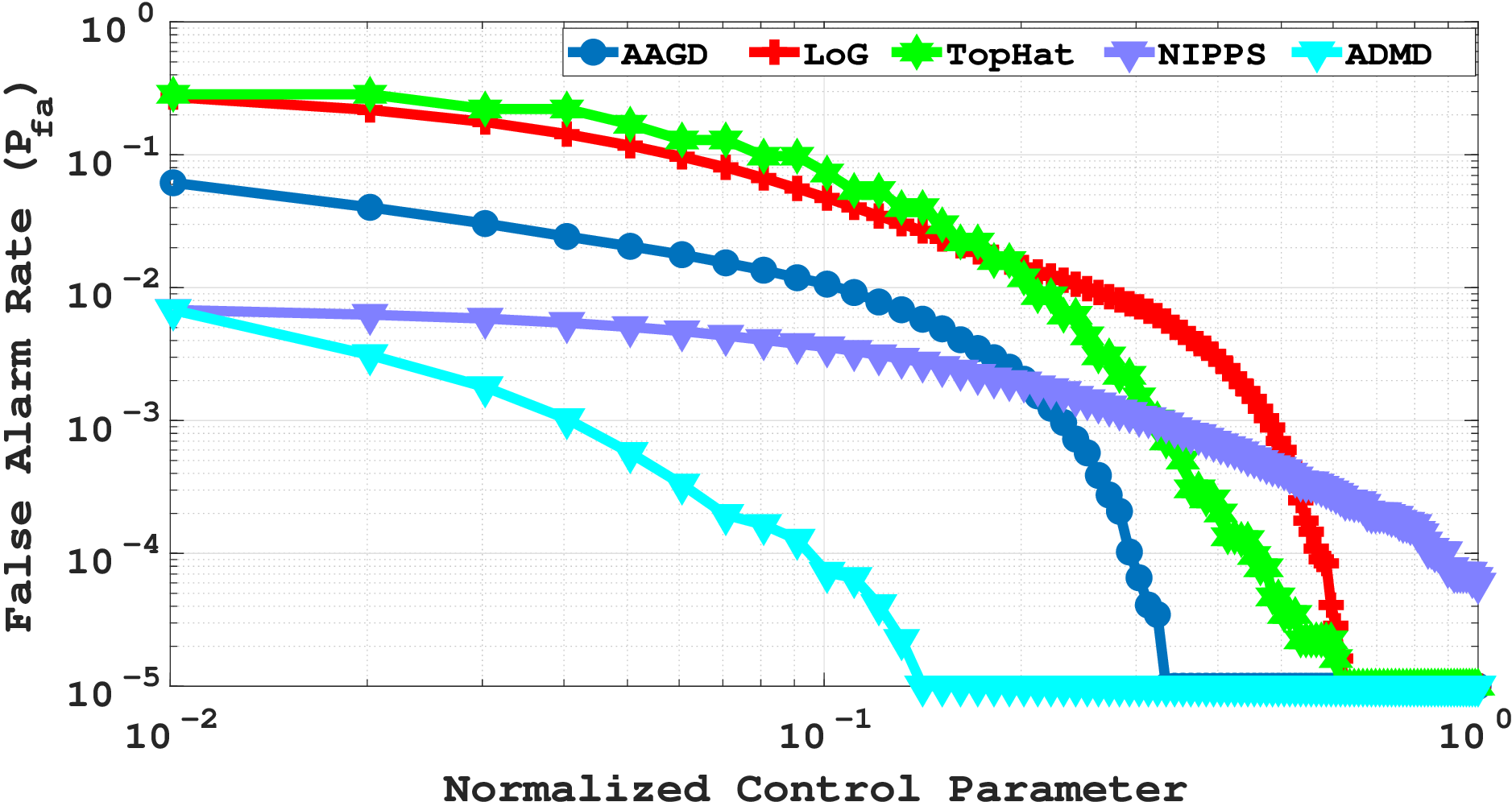}%
   		\label{fig:new_metric1}}
   	~
   	\subfloat[]{\includegraphics[width=3.3in]{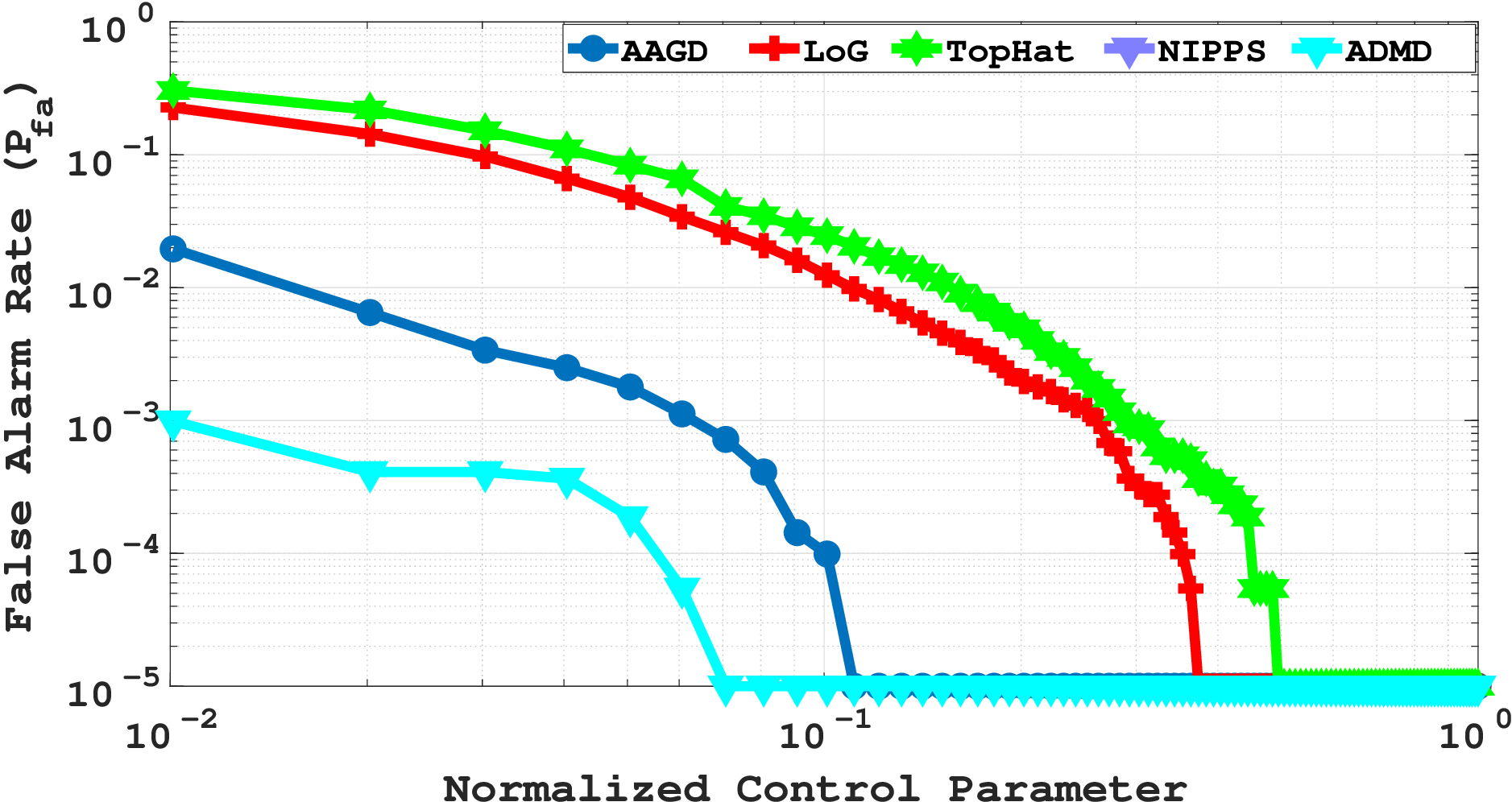}%
   		\label{fig:new_metric2}}
   	\\
   	\subfloat[]{\includegraphics[width=3.3in]{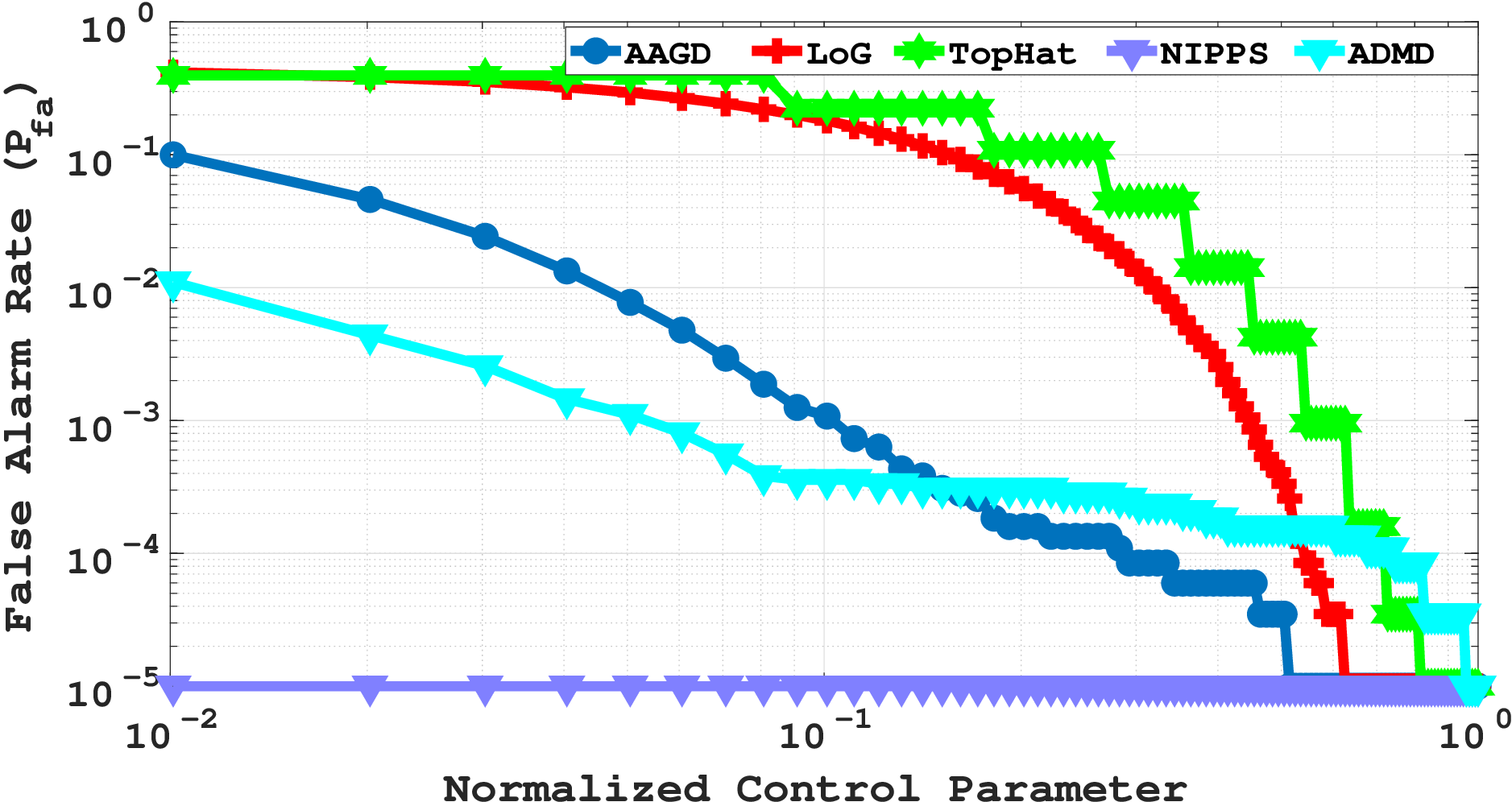}%
   		\label{fig:new_metric3}}
   	~
   	\subfloat[]{\includegraphics[width=3.3in]{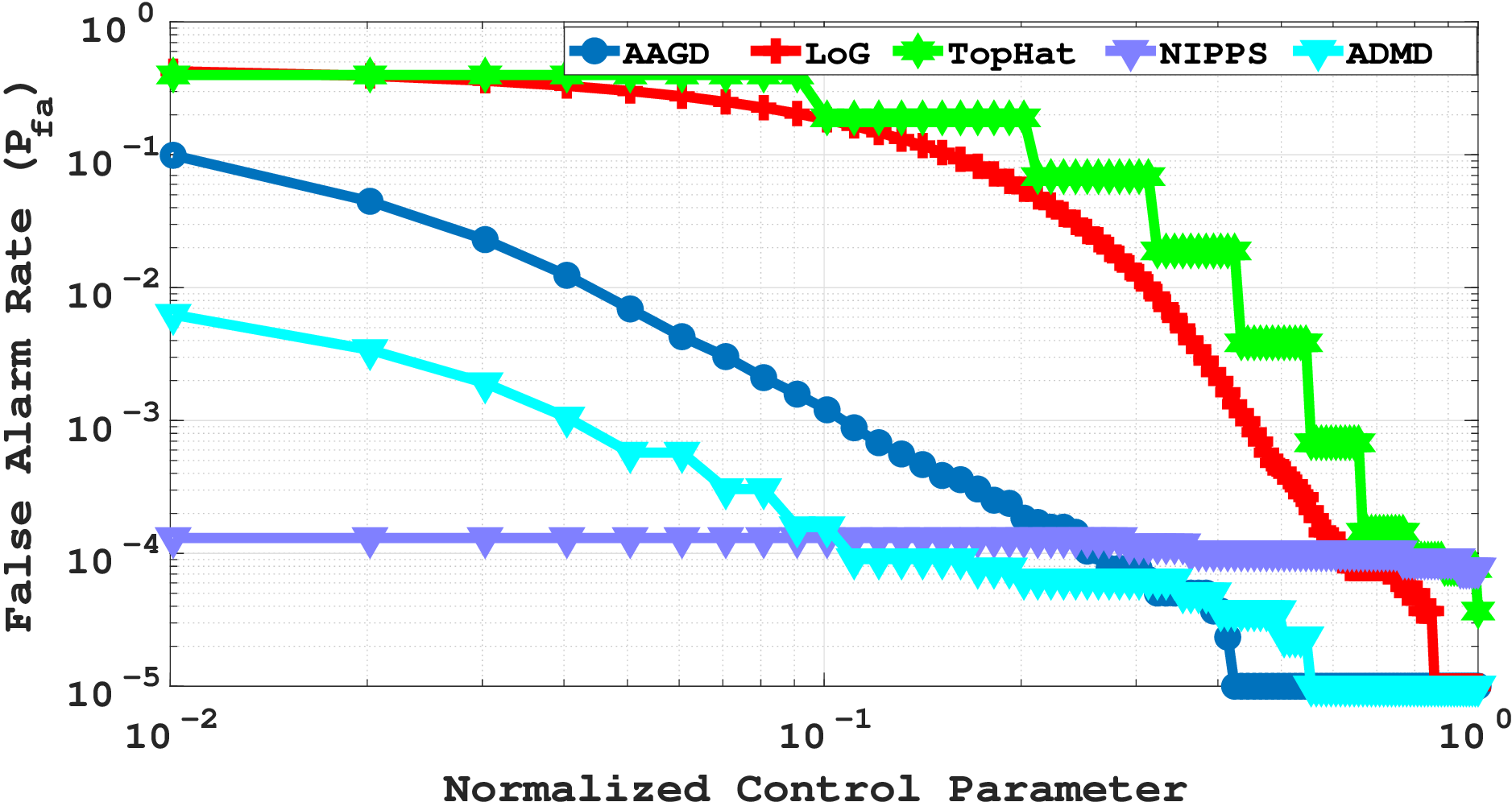}%
   		\label{fig:new_metric4}}
   	\\
   	\subfloat[]{\includegraphics[width=3.3in]{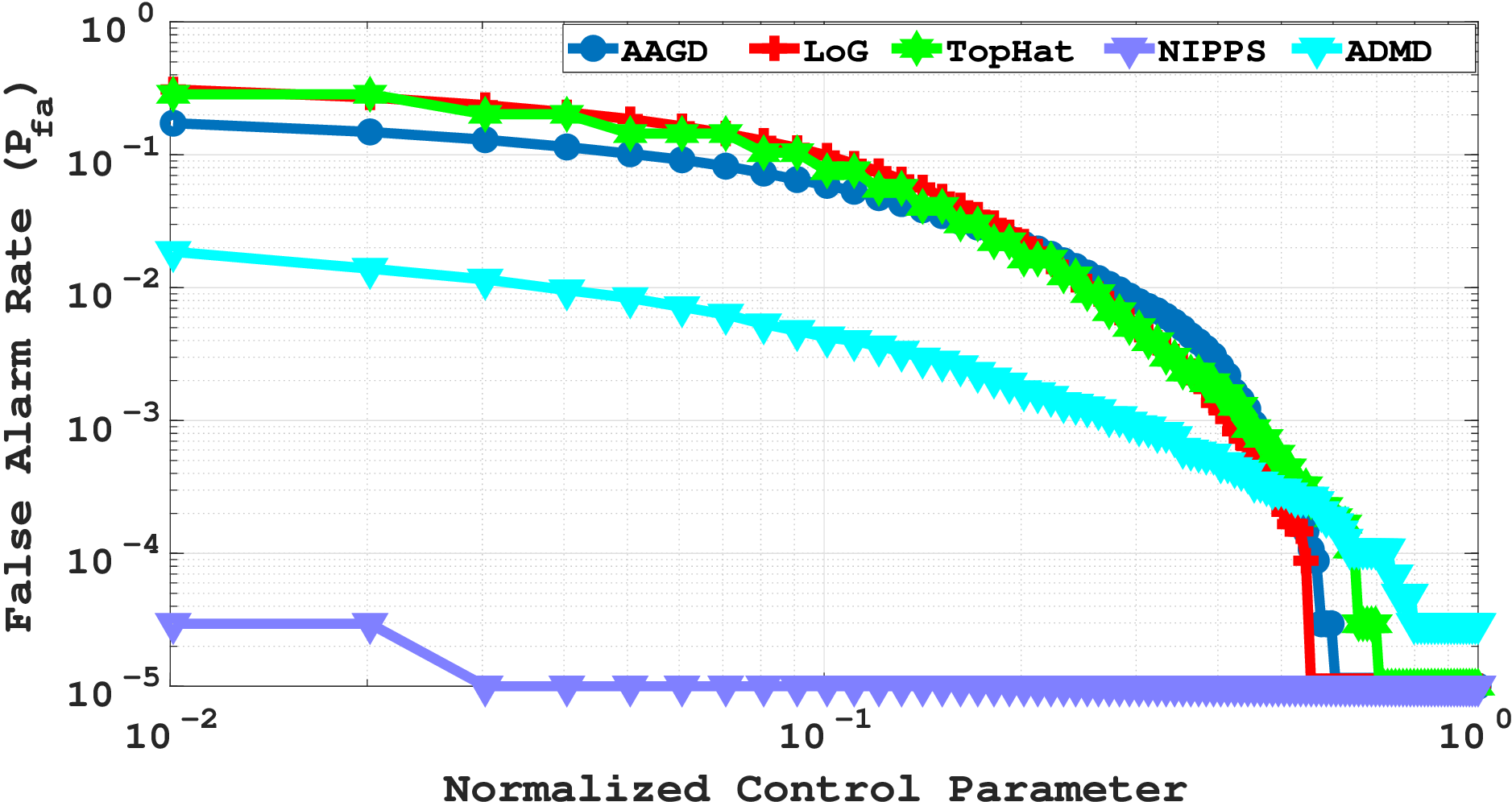}%
   		\label{fig:new_metric5}}
   		   	~
   	\subfloat[]{\includegraphics[width=3.3in]{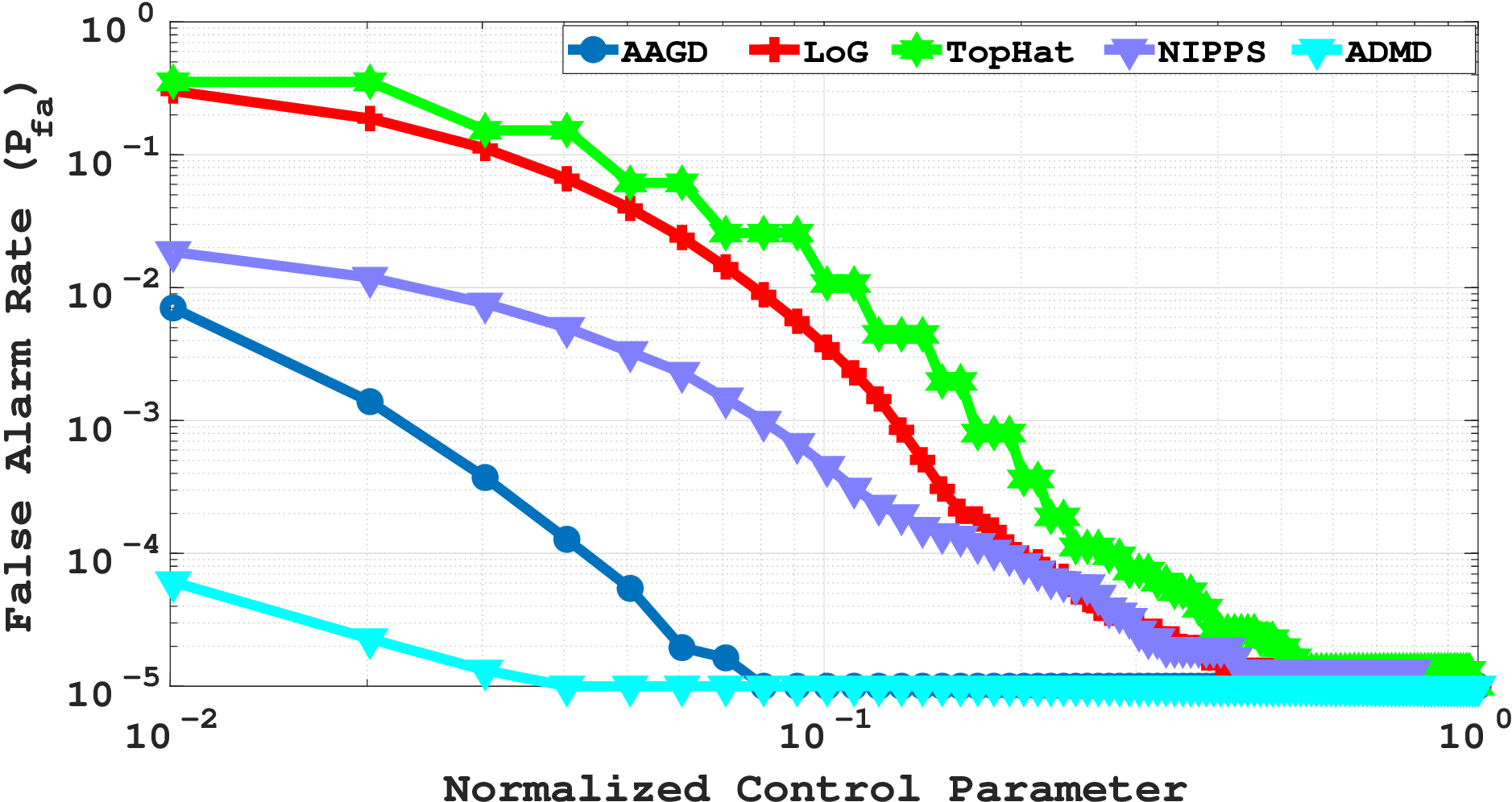}%
   		\label{fig:new_metric6}}
   	
   	\caption{The normalized ($P_{fa}$ -- $k$) curve in logarithmic scale. The detection performance characteristics curve for: a) the $1^{st}$ test image,   b) the $2^{nd}$ test image,    c) the $3^{rd}$ test image,     d) the $4^{th}$ test image,    e) the $5^{th}$ test image,     f) the $6^{th}$ test image.}
   	\label{fig:newmetrics}
   \end{figure}

\autoref{fig:newmetrics} shows the normalized ($P_{fa}$ -- $k$) curve to investigate the detection performance characteristics of different baseline algorithms, and fairly compare them. As expected, the LoG and TopHat algorithms shows higher false alarm rate across the entire plot. This is consistent with what one can observe from qualitative results. Again, AAGD algorithm shows moderate performance. Top algorithms are ADMD and NIPPS. ADMD stands on the first place for four scenarios, while NIPPS outperforms the other baselines in two scenarios. As shown in \autoref{fig:new_metric5}, the new metrics is completely consistent with the visual and qualitative  results (\autoref{fig:real_test}).

         \begin{figure}[t!]
   	\centering
   	   	\subfloat[]{\includegraphics[width=1.1in]{orig1.eps}%
   		\label{fig:new_thresh1}}
   	~
   	\subfloat[]{\includegraphics[width=1.1in]{aagd1.eps}%
   		\label{fig:new_thresh2}}
   	~
   	\subfloat[]{\includegraphics[width=1.1in]{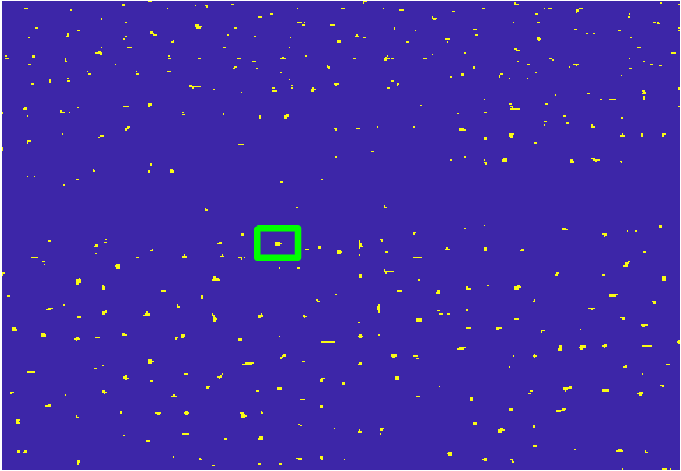}%
   		\label{fig:new_thresh3}}
   	~
   	\subfloat[]{\includegraphics[width=1.1in]{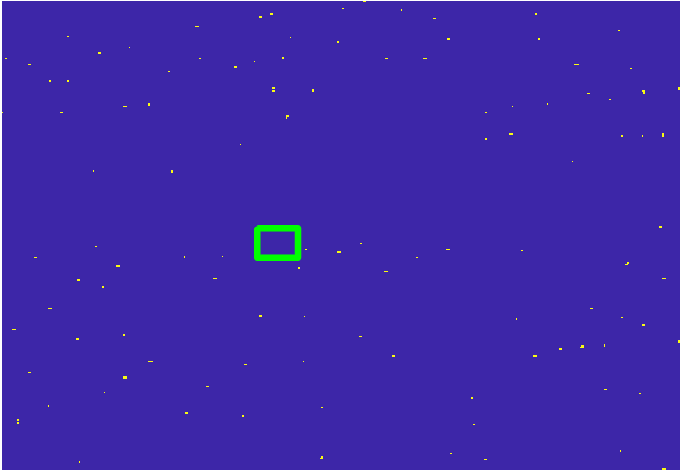}%
   		\label{fig:new_thresh4}}
   	~
   	\subfloat[]{\includegraphics[width=1.1in]{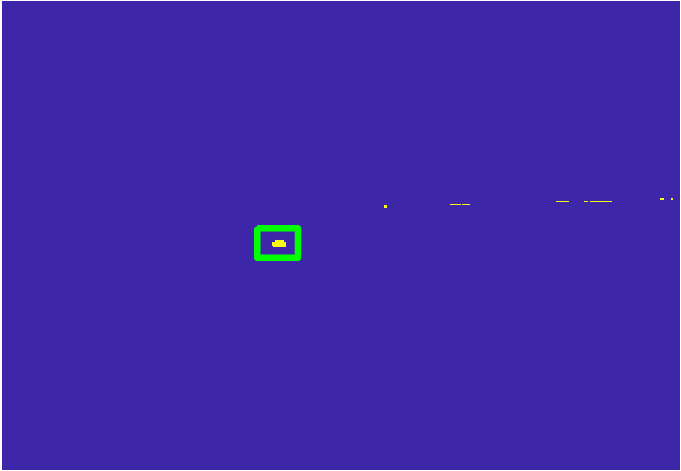}%
   		\label{fig:new_thresh5}}
   	~
   	\subfloat[]{\includegraphics[width=1.1in]{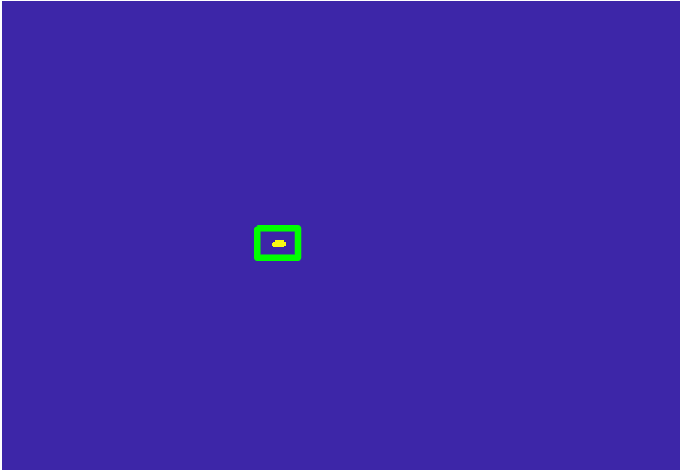}%
   		\label{fig:new_thresh6}}

   	\caption{Comparison of local statistics-based thresholding versus thresholding using \autoref{eq:adaptive_global_thresh}. a) Original infrared image, b) saliency map using AAGD algorithm, c) detection result using local thresholding $k_L=4$, d) detection result using local thresholding $k_L=6$, e) detection result using \autoref{eq:adaptive_global_thresh} $k_G=12$, and f) detection result using \autoref{eq:adaptive_global_thresh} $k_G=15$.}
   	\label{fig:threshcomp}
   \end{figure}
   
 Finally,  \autoref{fig:threshcomp} shows a comparison between local and global thresholds when are applied to a saliency map. As shown in the figure, regardless of the control parameter, the local one cannot eliminate background and extract true targets. However, the global one is completely successful in extracting target area.
\section{Conclusion}
 The development of new algorithms for infrared small target detection has gained significant attention in the past decade. However, many of these recent algorithms do not meet the requirements of practical applications, and there are limitations associated with common evaluation metrics. To gain a comprehensive understanding of effective evaluation metrics, it is essential to reveal the practical procedure of small target detection, with the thresholding operation playing a critical role. Without a proper thresholding strategy, previous advancements in target enhancement algorithm development would become obsolete. Empirical evidence has shown that local statistics-based thresholding is unsuitable for saliency map segmentation, while a global statistics-based threshold operation proves to be a better choice.

By incorporating global thresholding as the final step in the detection algorithm, modifications are made to enhance the signal-to-clutter ratio (SCR) metric's ability to reflect detection performance. Furthermore, three post-thresholding metrics are proposed to provide a comprehensive evaluation of different algorithms' performance. These improvements in both thresholding strategy and evaluation metrics contribute to the advancement of infrared small target detection techniques.

\bibliographystyle{IEEEtran}
\bibliography{ref.bib}
\end{document}